\documentclass[letterpaper, 10 pt, conference]{ieeeconf}  

\IEEEoverridecommandlockouts                              

\usepackage[letterpaper, left=0.735in, right=0.735in, bottom=0.750in, top=0.790in]{geometry}

\usepackage{multicol}
\usepackage{amsmath}
\usepackage{amsfonts}
\usepackage{algorithmic}
\usepackage{algorithm}
\usepackage{array}
\usepackage{float}

\usepackage{graphicx}
\usepackage[table]{xcolor}
\usepackage[dvipsnames,svgnames]{xcolor}
\usepackage{booktabs}
\usepackage{tabularx}
\usepackage{makecell}
\usepackage{colortbl}
\usepackage{multirow}
\usepackage{adjustbox}
\usepackage{subcaption}
\usepackage{csquotes}
\usepackage{cite}
\usepackage[bookmarks=true]{hyperref}
\usepackage{url}        
\usepackage{cleveref}
\usepackage{paralist}

\definecolor{lavender}{RGB}{210, 158, 182}
\definecolor{gnmBlue}{RGB}{41, 35, 219}
\definecolor{bridgerRed}{RGB}{219, 35, 35}

\newcommand{\prettygnm}{\textcolor{gnmBlue}{\textbf{GNM}}}
\newcommand{\prettyvint}{\textcolor{orange}{\textbf{ViNT}}}
\newcommand{\prettynomad}{\textcolor{lavender}{\textbf{NoMaD}}}
\newcommand{\prettybridger}{\textcolor{bridgerRed}{\textbf{NaviBridger}}}
\newcommand{\prettycross}{\textcolor{cyan}{\textbf{CrossFormer}}}
\newcommand{\prettyref}{\textcolor{Green}{\textbf{reference trajectory}}}

\title{\LARGE \bf
Can Vision Foundation Models Navigate? \\ Zero-Shot Real-World Evaluation and Lessons Learned 
}

\author{
\authorblockN{Maeva Guerrier, Karthik Soma, Jana Pavlasek, Giovanni Beltrame}
\authorblockA{Polytechnique Montreal\\
Email: maeva.guerrier@polymtl.ca}}

\begin{document}

\bstctlcite{IEEEexample:BSTcontrol}

\thispagestyle{empty}
\pagestyle{empty}


\makeatletter
\g@addto@macro\@maketitle{
  \begin{figure}[H]
  \setlength{\linewidth}{\textwidth}
  \setlength{\hsize}{\textwidth}
    \centering
    \includegraphics[width=0.95\linewidth]{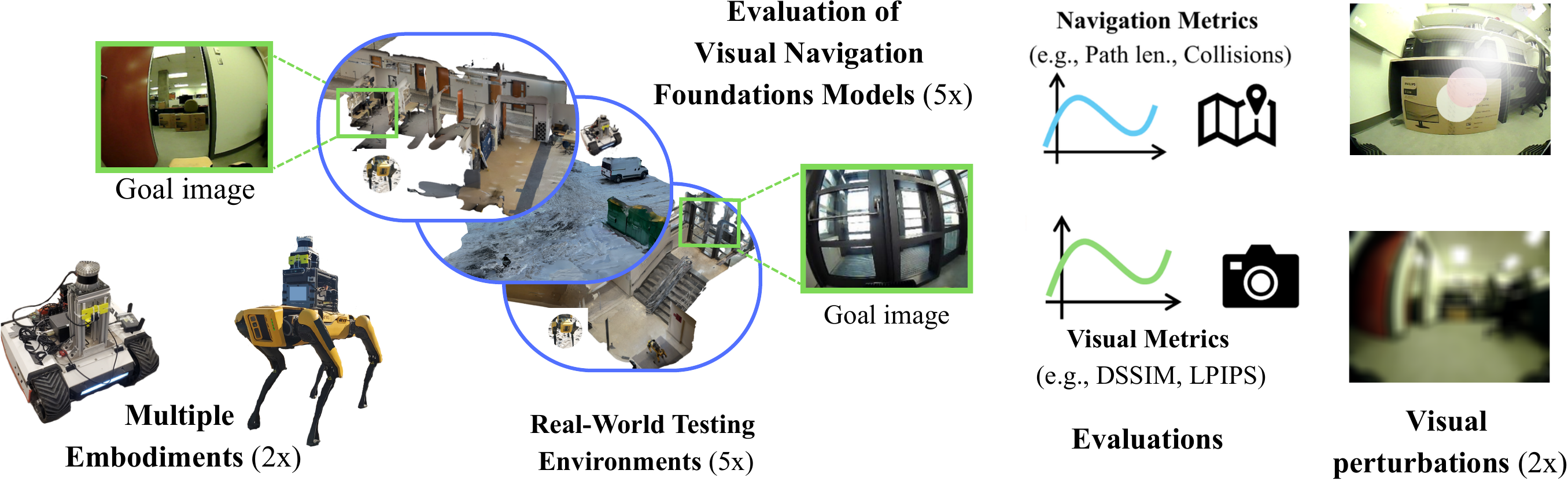}
      \caption{Visual Navigation Models promise to deliver generalizable navigation in novel environments. We provide a comprehensive analysis of 5 models with 2 robot embodiments over 5 real-world environments and propose metrics to better capture their performance.}
      \label{fig:pitch}
  \end{figure}
}
\makeatother

\maketitle


\begin{abstract}
Visual Navigation Models (VNMs) promise generalizable, robot navigation
by learning from large-scale visual demonstrations. Despite growing real-world
deployment, existing evaluations rely almost exclusively on success rate, whether
the robot reaches its goal, which conceals trajectory quality, collision behavior,
and robustness to environmental change.
We present a real-world evaluation of five state-of-the-art VNMs
(GNM, ViNT, NoMaD, NaviBridger, and CrossFormer) across two robot platforms and
five environments spanning indoor and outdoor settings. Beyond success rate, we
combine path-based metrics with vision-based goal-recognition scores and assess
robustness through controlled image perturbations (motion blur, sunflare).
Our analysis uncovers three systematic limitations: 
\begin{inparaenum}[(a)]
\item even architecturally
sophisticated diffusion- and transformer-based models exhibit frequent collisions,
indicating limited geometric understanding; 
\item models fail to discriminate
between different locations that are perceptually similar, however some semantics differences are present, causing goal prediction errors in repetitive environments; and 
\item performance degrades under distribution shift. 
\end{inparaenum}
We will publicly release our evaluation
codebase and dataset to facilitate reproducible benchmarking of VNMs.
\end{abstract}


\begin{figure*}[t]
    \centering
    \includegraphics[width=0.8\textwidth]{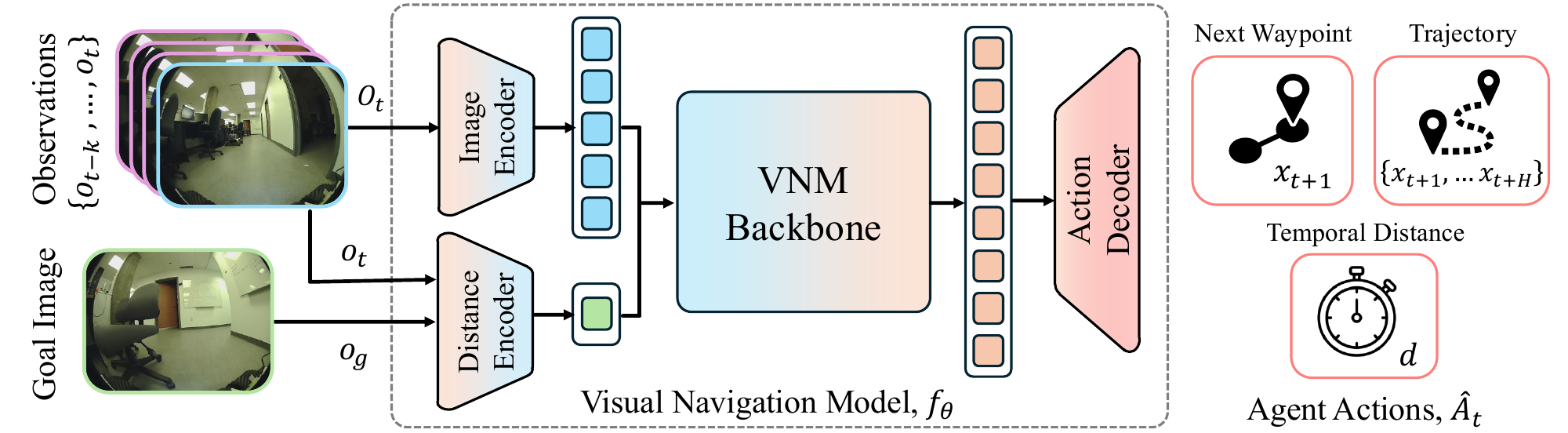}
    \caption{A Visual Navigation Model (VNM) takes as input a sequence of $k$ image observations $O_t$ and a goal image $o_g$ at timestep $t$. An image and distance encoder encode the observation sequence and the current ($o_t$) and goal observation respectively. The backbone and decoder vary by method. The model outputs action $\hat{A}_t$, which can take the form of a single next waypoint or a trajectory, and can also include the temporal distance $d$. }
    \label{fig:vnm_diagram}
\end{figure*}

\section{Introduction}



Visual navigation—guiding a robot to a goal using only camera images, without precomputed maps or expensive sensors—is a fundamental challenge in robotics. Traditional navigation methods rely on geometric maps that assume a static world, becoming unreliable as environments evolve and limiting their scalability to large-scale or dynamic settings.

Visual Navigation Models (VNMs)~\cite{gnm} learn a navigation policy from data,
allowing a robot to navigate toward a goal through visual observations alone. A
VNM takes as input a current observation image, a goal image, and a history of
observed images, then outputs navigation commands to drive the robot toward the
goal location. These models are trained on large robotic datasets~\cite{recon,
go_stanford, urmson2007tartan, karnan2022scand} and leverage visual cues to
navigate without precomputed maps.

These models have grown in popularity~\cite{vint,nomad,care}; however, we lack a comprehensive understanding of how these models actually perform in real-world deployment. Current evaluations rely almost exclusively on success rate metrics, typically defined by a distance threshold — this tells us whether a robot reached its goal, but reveals little about why models fail or how well they generalize across visually different environments. This evaluation gap is critical: foundation models for visual navigation are being deployed on real robots, from warehouse systems to NASA's Mars rover~\cite{NASA}, yet we cannot explain their failures or predict when they will struggle. This work directly addresses this gap.

This paper makes two primary contributions to the field of VNMs. First, we
introduce a comprehensive real-world evaluation of several state-of-the-art
VNMs. We consider GNM~\cite{gnm}, ViNT~\cite{vint}, NoMaD~\cite{nomad},
NaviBridger~\cite{bridger} and CrossFormer~\cite{crossformer}, across two
robotic platforms under three conditions: in familiar settings, controlled
distribution shift via image perturbations (motion blur, sunflare), and
out-of-distribution deployment. We will publicly release the evaluation dataset. The code can be found in  \href{https://github.com/MaevaGuerrier/vnm-zeroshot-eval}{Official Codebase}. Second,
through extensive real-world testing, we evaluate models using both traditional
path metrics and vision-specific metrics to assess goal understanding.


Our evaluation reveals three critical limitations in current VNMs. First, more
recent transformer-based and diffusion-based models consistently fail at
collision avoidance. This suggests that architectural complexity does not
guarantee geometric understanding, and that current training datasets lack
sufficient collision and recovery examples. Second, even when equipped with
low-level collision avoidance, VNMs suffer from goal prediction failures in
cluttered environments, frequently deviating from the intended trajectory because the models confuse
visually similar locations. Third, certain models performance degrade significantly under
distribution shift, highlighting limited robustness to environmental variations.
These findings demonstrate that while VNMs show promise for generalization,
substantial gaps remain before reliable real-world deployment is possible.

\section{Visual Navigation Models}

VNMs enable camera-only robot navigation using topological maps. They are trained to learn navigation with imitation learning on large-scale demonstration datasets and generalize to novel environments though learned visual features. We now introduce the problem formulation for the visual navigation problem and the particulars of the architectures of the VNMs considered in this work. 


Given a sequence of images $O_t = \{o_{t - k}, \dots, o_t\}$, consisting of a history of $k$ image observations, and a goal image $o_g$, a VNM $f_\theta$ generates an action:
\begin{equation}
    \hat{A}_t = f_\theta (O_t, o_g)
\end{equation}
The action $\hat{A}_t$ generally takes the form of a single next waypoint, $\hat{A}_t = x_{t+1}$, or a trajectory of waypoints over a horizon $H$, $\hat{A}_t = \{x_{t+1}, \dots, x_{t+H}\}$. Some models additionally include a ``temporal distance'' $d$ as an output which represents a unitless estimated measure of the distance from the current state $x_t$ to the goal. This process is visualized in Figure~\ref{fig:vnm_diagram}.

\textbf{Architecture.}
The typical architecture of a VNM is depicted in Figure~\ref{fig:vnm_diagram}. The image history sequence $O_t$ is encoded by an image encoder, while the current and goal images, $(o_t, o_g)$, are encoded by a distance encoder which represents the distance to the goal. The latent vectors from each model are concatenated and passed to a backbone then decoded into the action $\hat{A}_t$. The specific architectures of the encoders and backbone, as well as the action type, are dependent on the implementation. The specifics of the models considered in this work are described in Section~\ref{sec:vnm:variants}.

\textbf{Training.}
The models are trained via behavioral cloning on large-scale teleoperation datasets
The objective of the VNM is that subsequent applications of the model will result in a final state $x_T$ such that the observation $o_T$ approaches the goal image $o_g$. 
Specifically, $f_\theta$ is trained to imitate expert demonstrated actions $A_t$. Given a dataset of example actions,  their corresponding images, and the goal image, $\mathcal{D} = \{(O_i, o_{g,i}, A_i)\}_{i=1}^N$, the training objective is:
\begin{equation}
    \mathcal{L}(\theta) = \mathbb{E}_{\mathcal{D}} \left[~\text{dist}\left( f_\theta(O_i, o_{g, i}), A_i\right)~\right]
\end{equation}
Typically, the datasets $\mathcal{D}$ contains trajectories across diverse environments (e.g., 24 hours of office navigation data, approximately 35 hours of off-road data and 8 hours of sidewalks settings), using odometry as the ground truth action supervision.

\textbf{Topological Mapping.} 
In practice, for long-horizon trajectories where the goal is completely out of view from the initial position, determining the next action towards the goal image becomes very challenging. To mitigate this challenge, for deployment, a topological map $\mathbf{M} = (\mathcal{V}, \mathcal{E})$ is constructed from a set of nodes $\mathcal{V}$ and edges $\mathcal{E}$. The nodes $o_i\in\mathcal{V}$ are observations pre-collected in the environment, and the edges $e_{i,j}\in\mathcal{E}$ encode navigable connections between nodes. In contrast to metric maps, topological maps are sparse and do not have coordinates, making them inherently compact and memory-efficient~\cite{vnm_survey}.

During deployment, a closest-node belief is maintained at all times by selecting the node in 
$\mathbf{M}$ with the minimum predicted distance, estimated by the distance encoder of any VNM. A sequence of \textit{subgoals} is computed via graph search over the topological map, connecting the current closest node to the goal node. The VNM then navigates toward the goal by incrementally updating its active subgoal along this sequence. 

For efficient deployment, the nodes of a minimal topological map should, at a minimum, be placed at perceptually distinct locations to reduce ambiguity in closest-node belief estimation. Furthermore, since the vision encoders of the VNMs extract high-level visual features for navigation, it would be expected that VNMs exhibit robustness to minor environmental changes 
occurring after the construction of the topological map.

It is not necessary to rely on a pre-collected topological map for navigation. For instance, NoMaD~\cite{nomad} can performs undirected exploration through goal masking, without requiring any topological map. Furthermore, subgoal images can be generated using learned generative processes or world models~\cite{navworldmodel} conditioned on the current observation, for directed or undirected navigation.

\subsection{Architecture Variants}\label{sec:vnm:variants}


\begin{table}[b]
\renewcommand{\arraystretch}{1.}
    \centering
    \caption{The specific architectures of the Visual Navigation Models (VNMs) considered in this paper. Methods differ in their backbone architecture and output type. Some models introduce specific architecture choices within the VNM pipeline.
    }
    \resizebox{\linewidth}{!}{%
    \begin{tabular}{l p{0.17\columnwidth} p{0.19\columnwidth} c}
        \toprule
        \textbf{Model} & \textbf{Backbone} & \textbf{Specifics} & \textbf{Output} \\
        \midrule
        \prettygnm{}~\cite{gnm} & Fully Connected & Shared action space & $(\{x_i\}_{t+1}^{t+H}, d)$ \\
        \prettyvint{}~\cite{vint} & Transformer  & Goal Fusion & $(x_t, d)$\\
        \prettynomad{}~\cite{nomad} & Diffusion &  Goal Masking & $\{x_i\}_{t+1}^{t+H}$\\
        \prettybridger{}~\cite{bridger} & Diffusion & Learned action prior & $\{x_i\}_{t+1}^{t+H}$ \\
        \prettycross{}~\cite{crossformer} & Transformer & Embodiment specific head & $x_t$ \\
        \bottomrule
    \end{tabular}
    }
    \label{tab:vmn_arch}
\end{table}

\begin{table*}[htb]
    \captionsetup{width=\columnwidth}
    \centering
    \small
    \caption{Path length (\textit{p.len}) and Goal Distance (\textit{dist.}) across environments and models (mean$\pm$standard-deviation).}
    \begin{tabular}{l*{2}{c}*{2}{c}}
    & \multicolumn{2}{c}{\textbf{Corridor}} 
    & \multicolumn{2}{c}{\textbf{Office loop}} \\
    \cmidrule(lr){2-3} \cmidrule(lr){4-5}
    \textbf{Method} & dist. & p.len. & dist. & p.len. \\
    \midrule
    \rowcolor{gray!15}
    \prettygnm{} & 0.33$\pm$0.13 & 4.52$\pm$0.16 & 9.49$\pm$9.43 & 19.28$\pm$10.03 \\
    \rowcolor{gray!5}
    \prettyvint{} & 0.40$\pm$0.09 & 4.81$\pm$0.14 & 3.53$\pm$2.19& 32.51$\pm$2.57  \\
    \rowcolor{orange!15}
    \prettynomad{} & 0.56$\pm$0.09 & 5.25$\pm$0.15 & 7.22$\pm$5.80 & 30.04$\pm$9.45 \\
    \rowcolor{orange!5}
    \prettybridger{}  & 0.63$\pm$0.16 & 4.46$\pm$0.63 & 17.19$\pm$5.29 & 10.66$\pm$5.61  \\
    \prettycross{}  & 1.73$\pm$0.00 & 2.34$\pm$0.04 & - & - \\
    \bottomrule
    \end{tabular} 

    \begin{tabular}{l*{2}{c}*{2}{c}*{2}{c}}
    & \multicolumn{2}{c}{\textbf{Arena}} 
    & \multicolumn{2}{c}{\textbf{Stairs}} 
    & \multicolumn{2}{c}{\textbf{Snow}} \\
    \cmidrule(lr){2-3} \cmidrule(lr){4-5} \cmidrule(lr){6-7}
    \textbf{Method} & dist. & p.len. & dist. & p.len. & dist. & p.len. \\
    \midrule
    \rowcolor{gray!15}
    \prettygnm{} & 0.91$\pm$1.37 & 23.82$\pm$5.79 & 1.00$\pm$0.28 & 6.41$\pm$0.19 & 1.12$\pm$0.28 & 19.76$\pm$0.20 \\
    \rowcolor{gray!5}
    \prettyvint{} & 3.09$\pm$0.50 & 28.02$\pm$6.22 & 0.35$\pm$0.30 & 6.48$\pm$0.17 & 9.96$\pm$6.42 & 12.59$\pm$9.92 \\
    \rowcolor{orange!15}
    \prettynomad{} & 7.97$\pm$3.90 & 22.50$\pm$6.17 & 0.46$\pm$0.36 & 6.59$\pm$0.27 & 10.76$\pm$3.74 & 9.85$\pm$4.66 \\
    \rowcolor{orange!5}
    \prettybridger{}  & 11.23$\pm$3.14 & 31.46$\pm$8.91 & 1.50$\pm$1.91 & 12.91$\pm$5.79 & 4.78$\pm$0.76 & 16.67$\pm$0.31 \\
    \prettycross{}  & 10.56$\pm$3.61 & 28.67 $\pm$ 3.10 & - & - & - & - \\
    \bottomrule
    \end{tabular}
    \label{tab:precision_metrics_all_envs}
\end{table*}

A summary of the major architecture differences for the methods considered in this paper can be found in Table~\ref{tab:vmn_arch}. Models vary in their choice of image and distance encoders, as well as their output representation.
\prettygnm{}~\cite{gnm} uses MobileNetv2~\cite{dong2020mobilenetv2} for both image and distance encoding, \prettyvint{}~\cite{vint} and \prettynomad{}~\cite{nomad} use EfficientNet-B0~\cite{tan2019efficientnet} for both, \prettybridger{}~\cite{bridger} uses a dual-encoder Transformer for both, and \prettycross{}~\cite{crossformer} uses a ResNet-26 encoder~\cite{he2016deep} for  images encoding with ViNT handling distance prediction. 

GNM introduces a shared action space across robot embodiments to enable zero-shot deployment. ViNT builds on GNM by replacing GNM's fully connected layers with a Transformer decoder, introducing a goal token that fuses the current observation and goal image enabling learning joint features between the two. NoMaD extends ViNT by incorporating a diffusion process for multi-modal action prediction, as well as goal masking to support both directed and exploratory navigation. NaviBridger builds on NoMaD and incorporates a learned action prior using a conditional variational autoencoder. CrossFormer introduces embodiment specific transformer heads to improve scalability across diverse robot types.

\section{Related work}


Recent generalist robotics policies demonstrate strong performance across manipulation and navigation tasks on diverse embodiments~\cite{brohan2023rt1roboticstransformerrealworld, octo_2023}. These policies train on large-scale robotic datasets~\cite{open_x, walke2024bridgedatav2datasetrobot} to enable zero-shot transfer. Transformer architectures have become standard for processing these datasets. CrossFormer~\cite{crossformer} uses modular attention to control multiple embodiments including arms, quadrupeds, wheeled robots, and drones. This transformer-based approach has been extended to navigation applications~\cite{vint}.

Early learning-based navigation methods used semantic cues and self-supervised learning but struggled with long-horizon trajectories~\cite{NEURIPS2020_2cd4e8a2, radosavovic2022realworldrobotlearningmasked}. Topological graph representations address this limitation for RGB-based navigation~\cite{gnm, dgmem, selfi}. ViNT~\cite{vint} claims to achieve long-horizon navigation using topological maps and transformer encoders trained on large datasets~\cite{go_stanford, recon, urmson2007tartan, karnan2022scand}, enabling zero-shot generalization across embodiments and environments.

Recent models combine diffusion methods with transformers~\cite{nomad} to generate waypoints conditioned on RGB observations. Diffusion policies enable multi-modal behaviors by modeling complex action distributions. NaviBridger~\cite{bridger} refines this approach by incorporating action priors through diffusion bridges. A diffusion bridge constrains the diffusion process between two fixed distributions, to steer the starting distribution to a specified endpoint.


\textbf{Classical Approaches in VNMs.}
A recent trend integrates classical control into VNMs though scene-specific cost. Specifically, NaviDiffusor~\cite{zeng2025navidiffusorcostguideddiffusionmodel} introduces \textit{cost-guided sampling} to guide the reverse diffusion process toward collision-free paths. Similarly, CARE~\cite{care} adopt a reactive planning strategy, using a repulsive module to adjust paths at runtime.


\section{Beyond Success Rate: The Need for Comprehensive Robotics Metrics in Visual Navigation}

Visual navigation models are predominantly evaluated using success rate (SR), defined as fractional progress toward the goal~\cite{vint} or proximity to the closest node in a topological goal map~\cite{crossformer}. While SR provides a simple, comparable benchmark, it obscures trajectory quality, collision frequency, localization precision, and navigation efficiency. Critically, SR implies a fixed threshold for success, whereas real-world deployment requirements vary by application.

Traditional robotic navigation research employs richer metrics, such as collision rate, path length, smoothness, and final distance to goal, that reveal performance dimensions SR cannot capture. Two trajectories with identical SR may differ drastically in collision frequency or deviation from reference paths, yet current VNM evaluations~\cite{vint, gnm, crossformer} provide limited insight into these failure modes.

Vision-based navigation introduces additional evaluation challenges absent in classical approaches. Unlike LiDAR or depth-based systems, VNMs are sensitive to lighting changes, viewpoint variation, and scene appearance~\cite{vnm_survey}. 


Comprehensive evaluation of VNMs therefore requires metrics spanning three domains: path quality, visual perception, and robustness to distribution shift. We evaluate state-of-the-art VNMs across two robotic platforms under familiar and unseen conditions using metrics drawn from all three domains.



\section{Dataset \& Evaluation Setup}
\begin{figure}[tb]
  \centering
  \includegraphics[width=0.75\columnwidth]{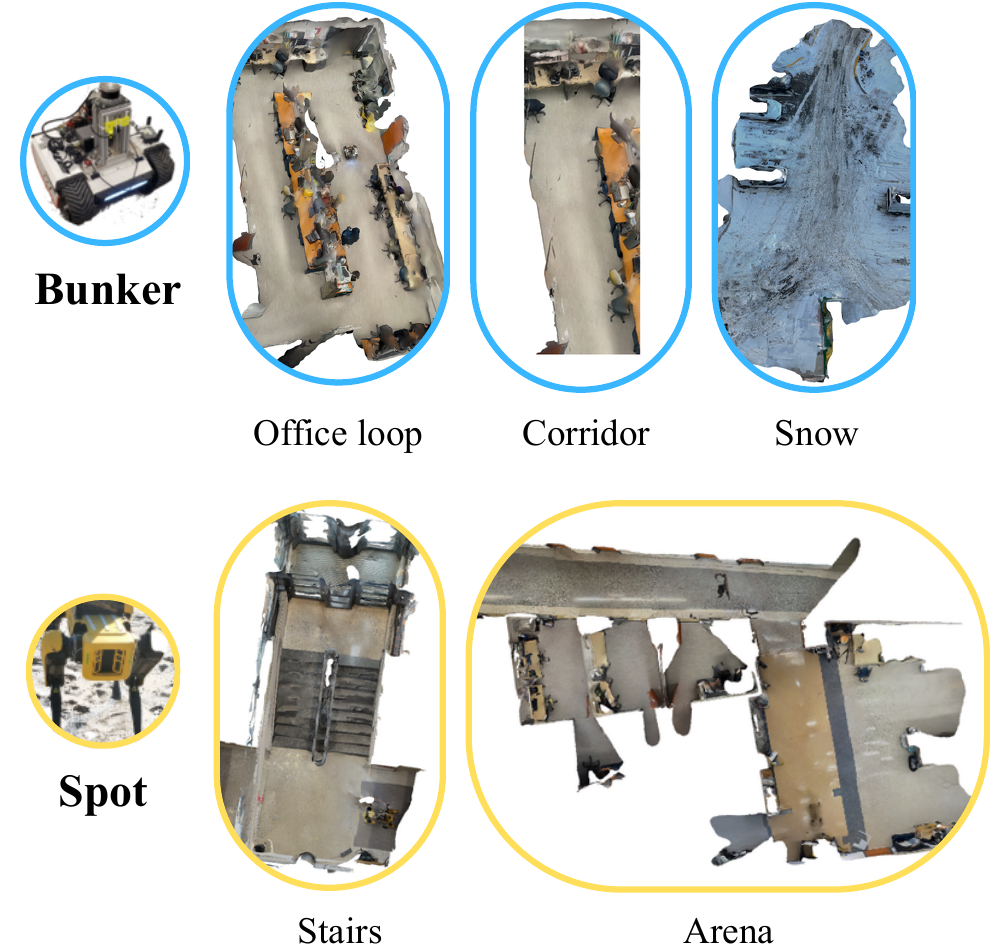}
  \caption{Real-world evaluation environments (indoor and outdoor). 
  Blue panels correspond to rover while yellow are quadruped deployments. 
  }
  \label{fig:meet_the_envs}
\end{figure}

We evaluate five real-world environments (see Figure~\ref{fig:meet_the_envs}) of varying difficulty, from familiar indoor office settings (e.g., desks, chairs) to out-of-distribution snowy terrain, deployed on quadruped (\textit{Spot}) and a tracked mobile robot (\textit{Bunker}). We evaluate the action-enriched model (CrossFormer) in two settings: a simple environment and a visually ambiguous one. This isolates whether this range of models improves the performance of VNMs in both easy and challenging conditions.


\textbf{Evaluation Metrics.} We evaluate models across two categories. \textit{Vision-based metrics} compare the final observation against the goal image using LPIPS~\cite{lpips}, DSSIM~\cite{fu2023dreamsim} (\textit{lower is better for both}), and PSNR~\cite{psnr} (\textit{higher is better}). \textit{Navigation metrics} include path length, distance to goal, collision occurrence, goal prediction and topological node error. The topological node error is computed by summing the absolute difference between the model's closest node belief and the ground truth node at each trajectory position, then averaged.
For the remaining distance to goal, we compute: (1) checkpoints, equally spaced along the reference trajectory, (2) the remaining distance to goal is computed from the last visited checkpoint along the remaining distance from that checkpoint to the final goal position by summing up the segment. If the last checkpoint is the goal, the Euclidean distance is computed directly from that position. The collision occurrence marks a deployment as collision if the robot collides at any point, immediately terminating the deployment. Since the trial ends upon first collision contact, at most one collision is recorded per deployment. Goal prediction indicates when the model believes it has reached the goal, regardless of its actual position. It does not imply physical goal achievement and must be interpreted alongside the remaining distance to goal.

\subsection{Meet The Environments}

\textbf{Corridor:} A short-horizon, straight-line trajectory (\textit{3.749m, Bunker}) toward a cardboard box (\textit{see Figure~\ref{fig:corridor_figure}}). 

\begin{figure}[tbp]
    \centering
       \begin{tabular}{@{}c@{\hspace{0.5em}}c@{}}
            \includegraphics[width=0.38\columnwidth]{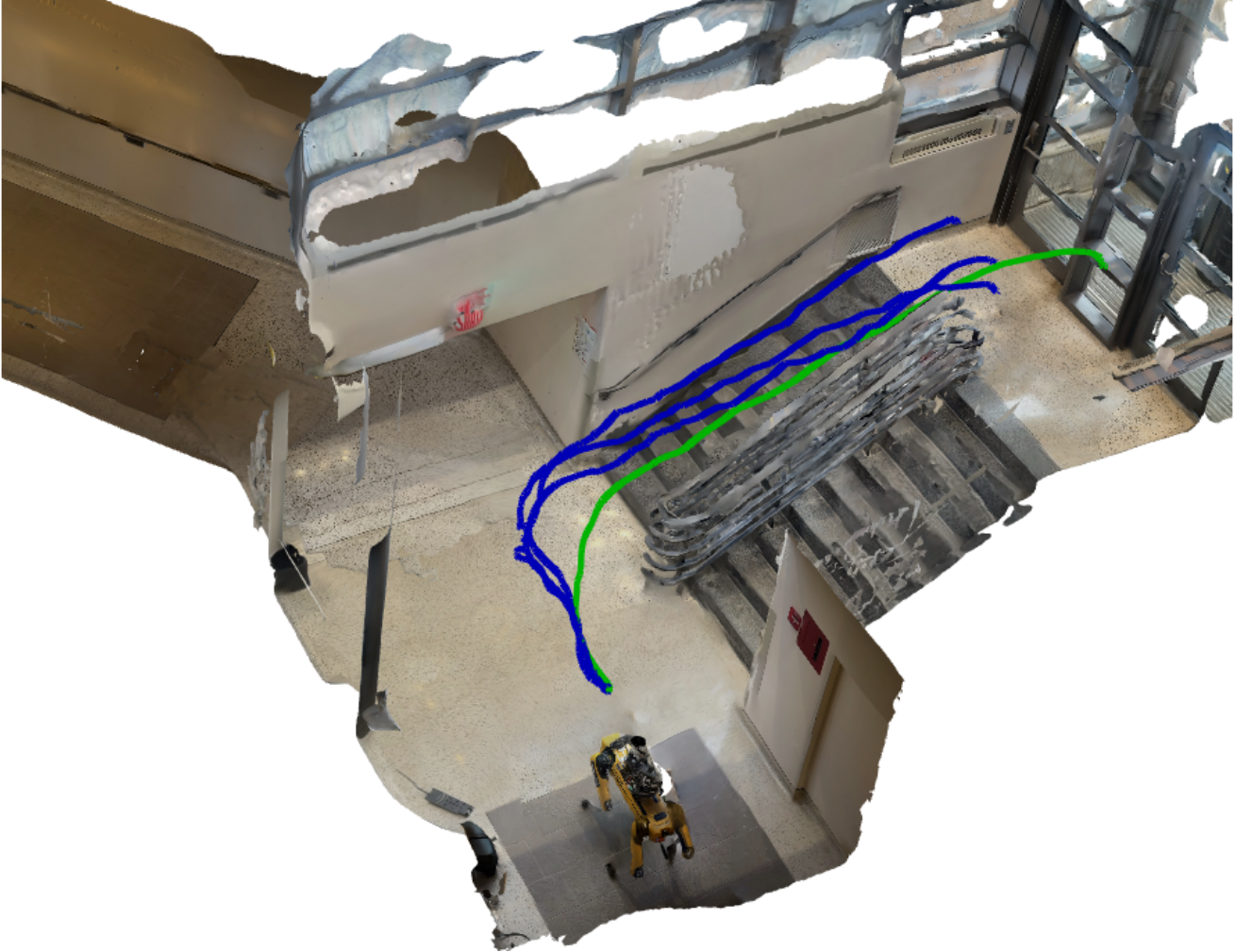} &
            \includegraphics[width=0.38\columnwidth]{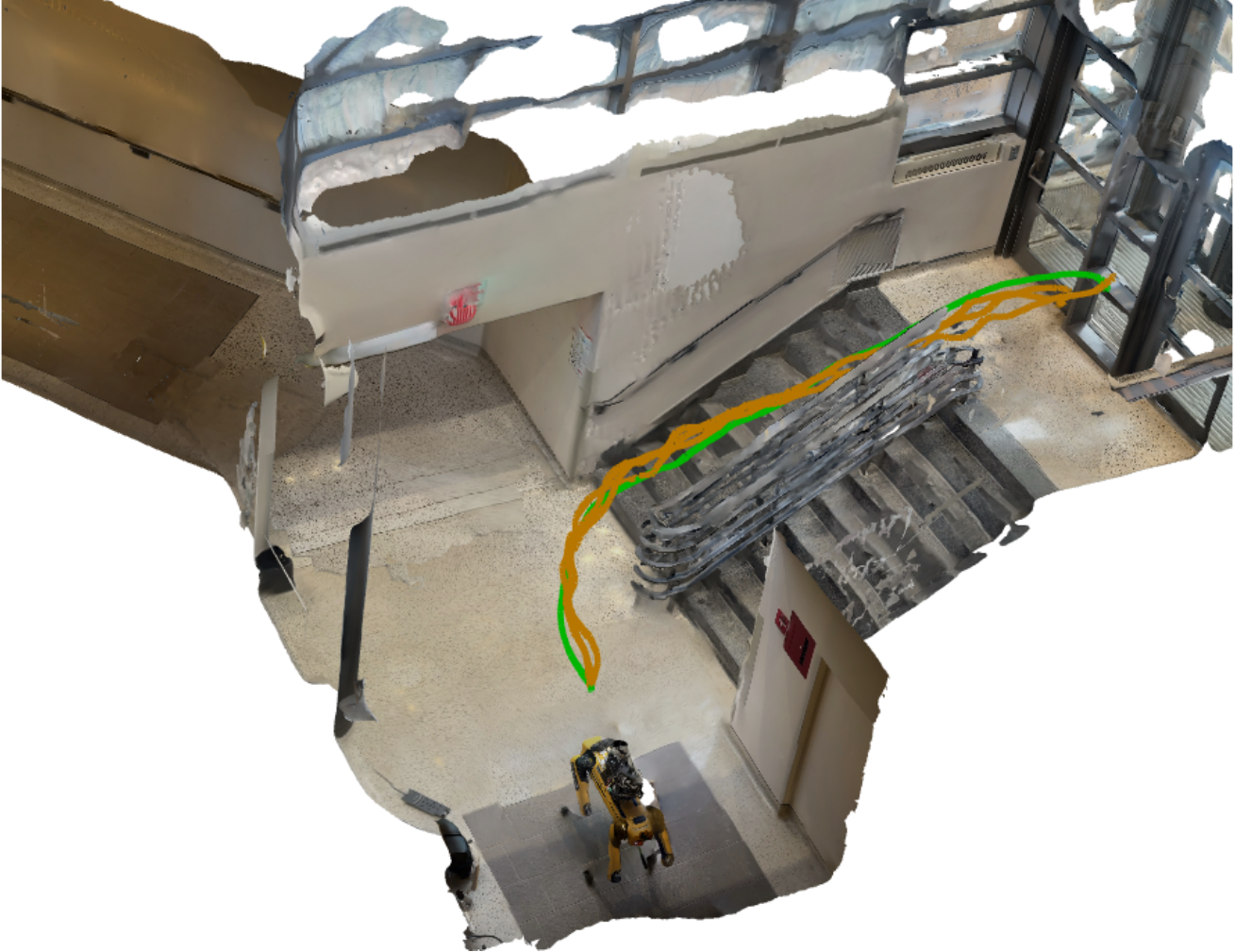} \\
            \includegraphics[width=0.38\columnwidth]{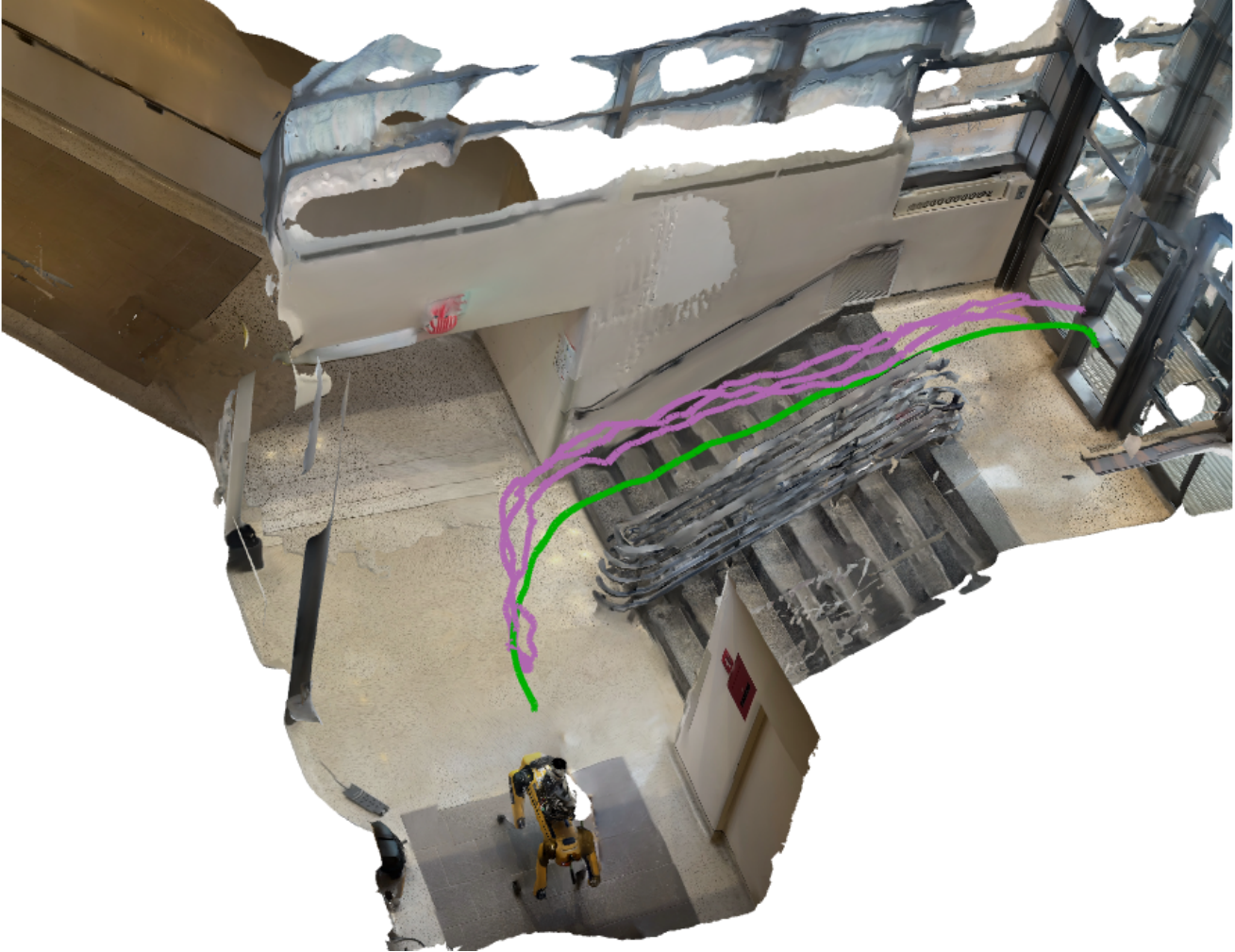} &
            \includegraphics[width=0.38\columnwidth]{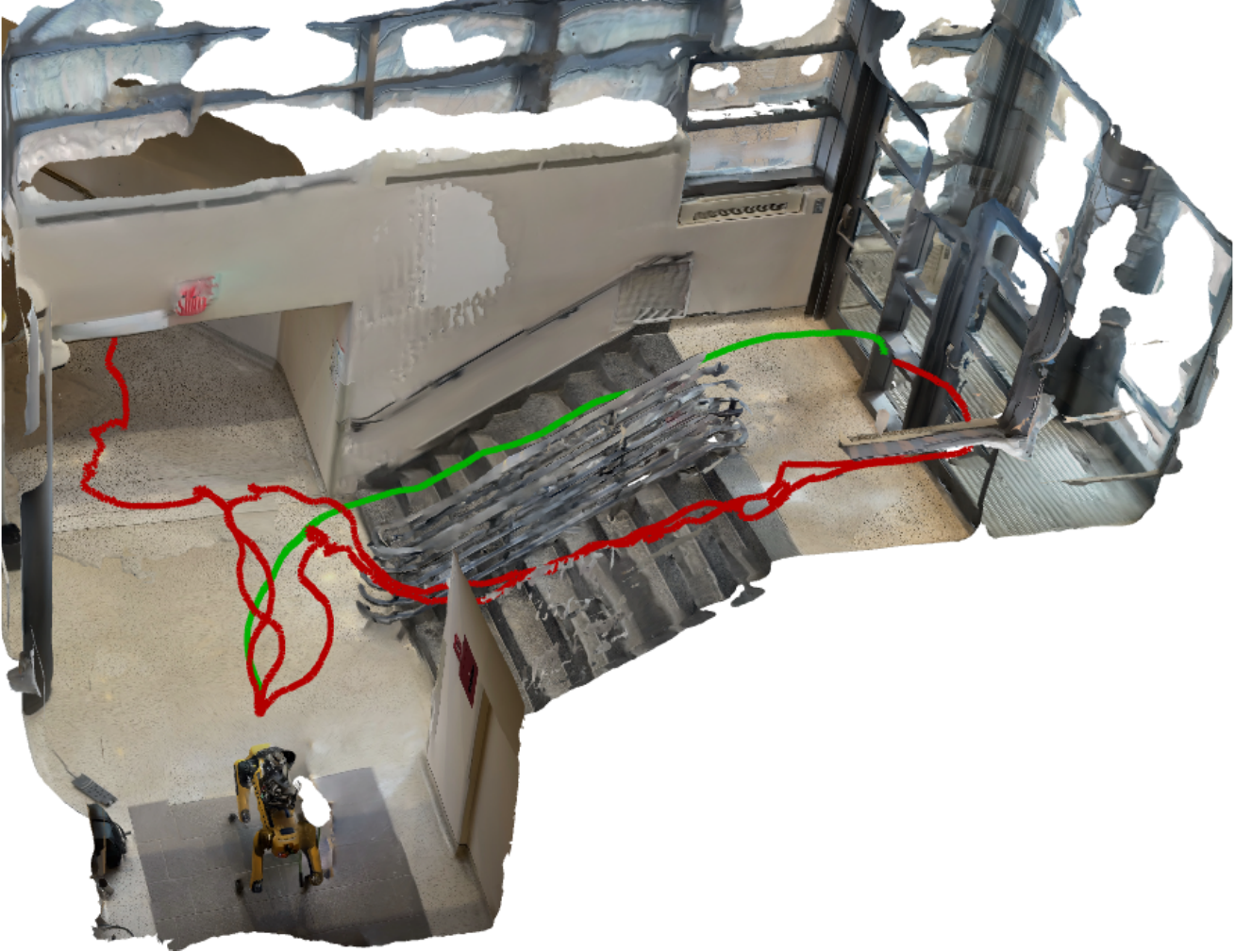} \\
        \end{tabular}
    \caption{All \textbf{Stairs} trajectories for \prettygnm{}, \prettyvint{}, \prettynomad{} and \prettybridger{} (see Table~\ref{tab:generalization_metrics_all_env}) with the \prettyref{}.}
    \label{fig:stairs_mesh}
\end{figure}

\textbf{Stairs:} An ascending staircase trajectory (\textit{6.247m, Spot}) featuring two visually identical staircases, testing robustness to perceptually similar features and generalization (\textit{see Figure~\ref{fig:stairs_mesh}}). 

\begin{figure}[htbp]
    \centering
       \begin{tabular}{@{}c@{\hspace{0.5em}}c@{}}
            \includegraphics[width=0.30\columnwidth]{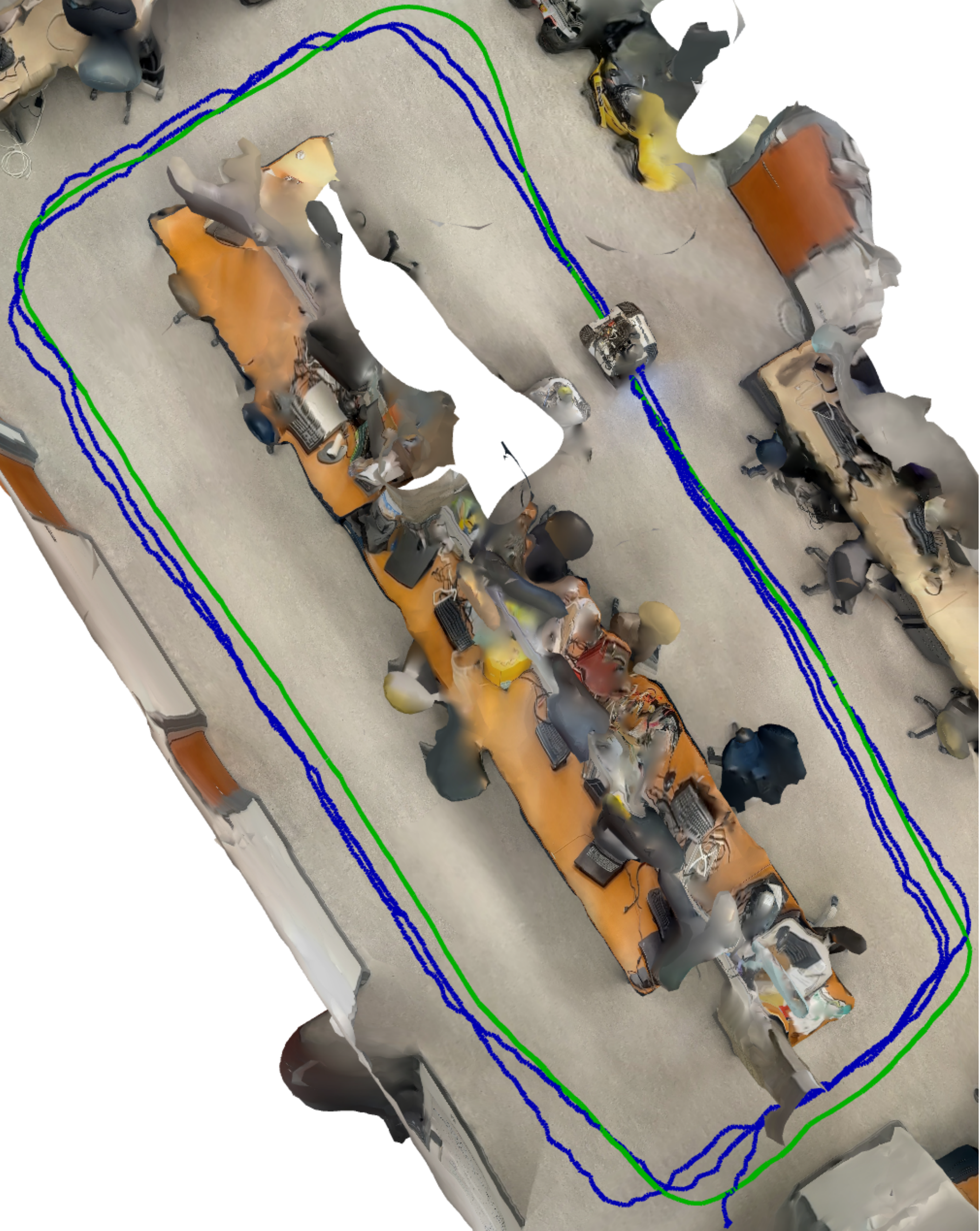} &
            \includegraphics[width=0.30\columnwidth]{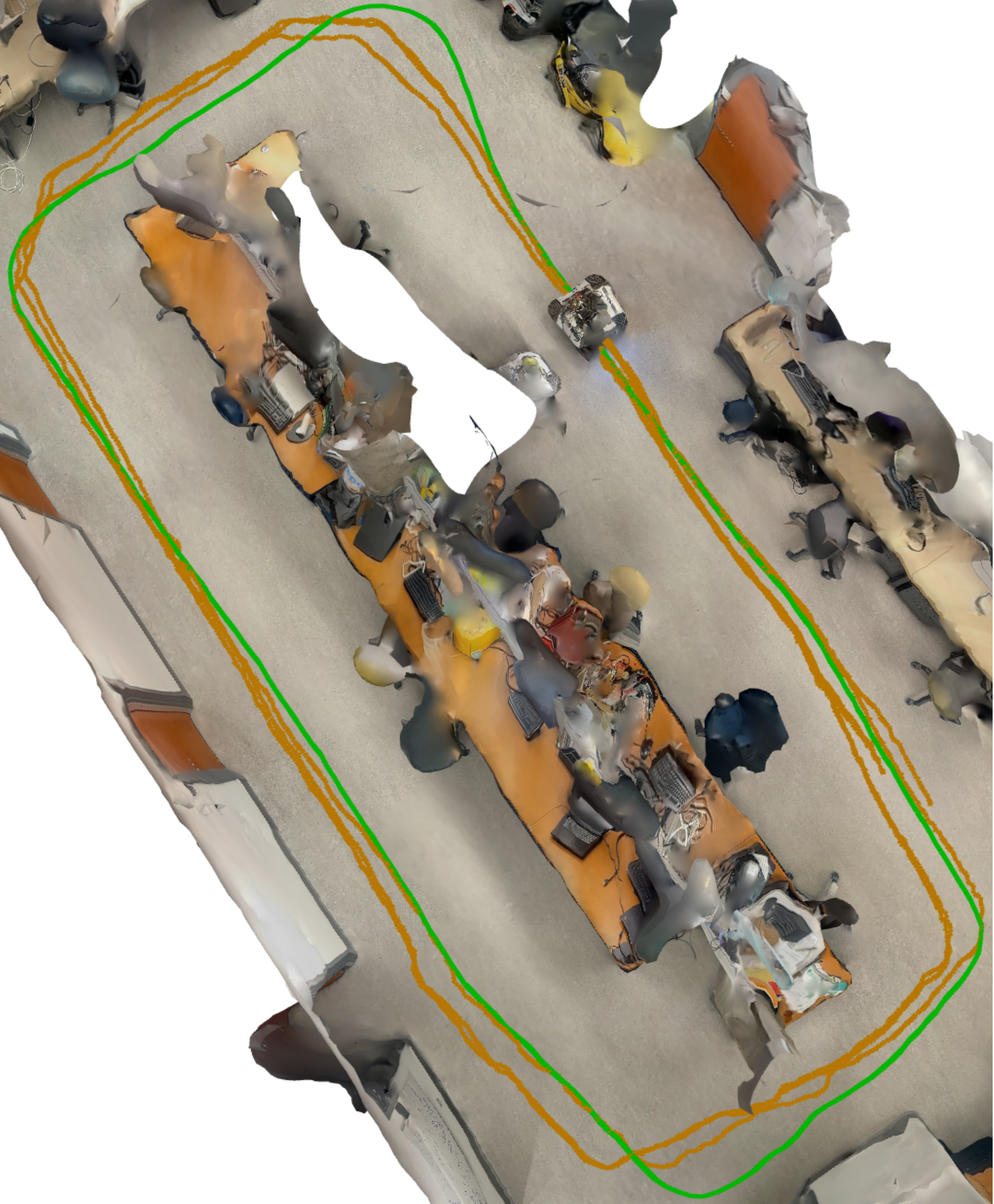} \\
            \includegraphics[width=0.30\columnwidth]{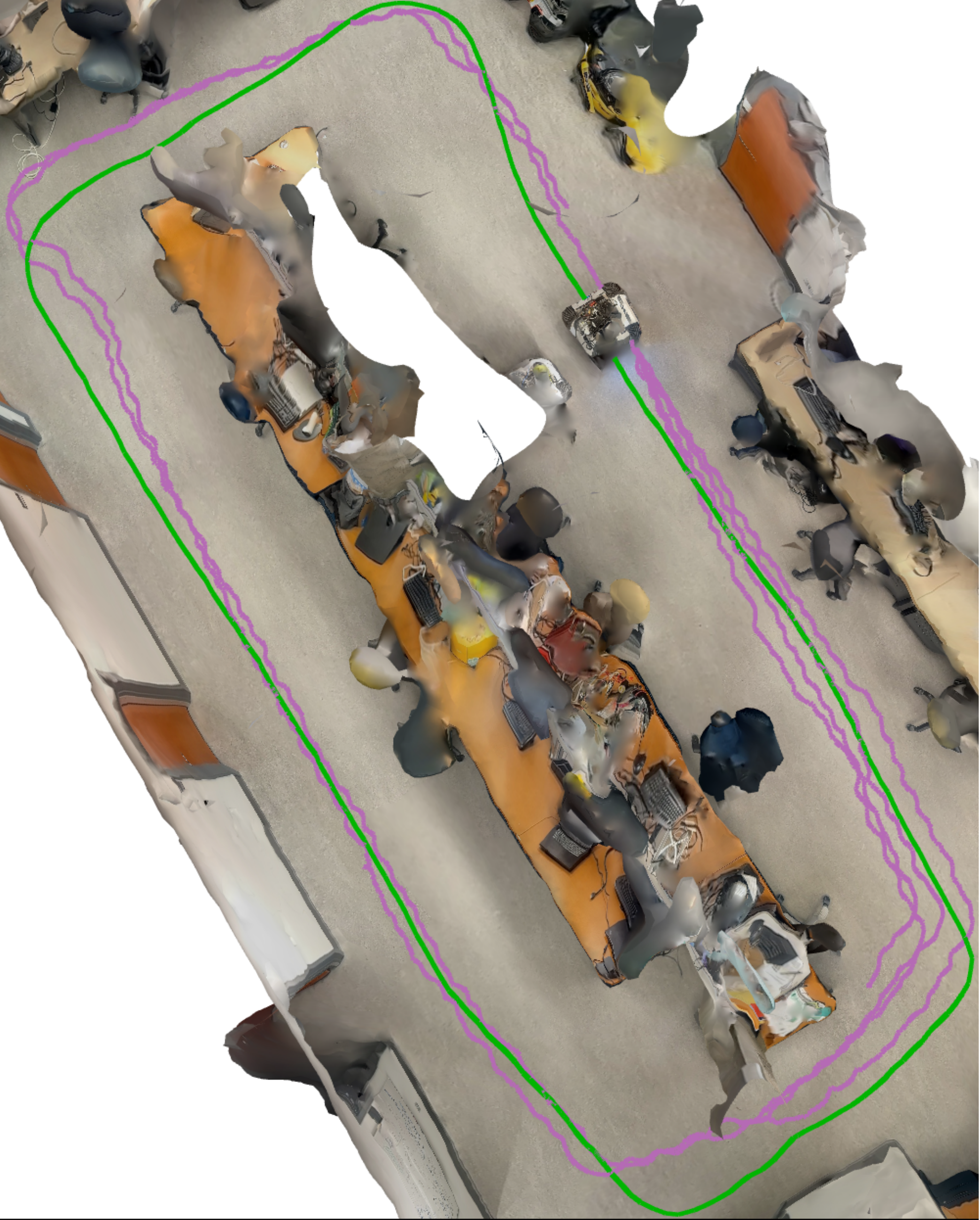} &
            \includegraphics[width=0.30\columnwidth]{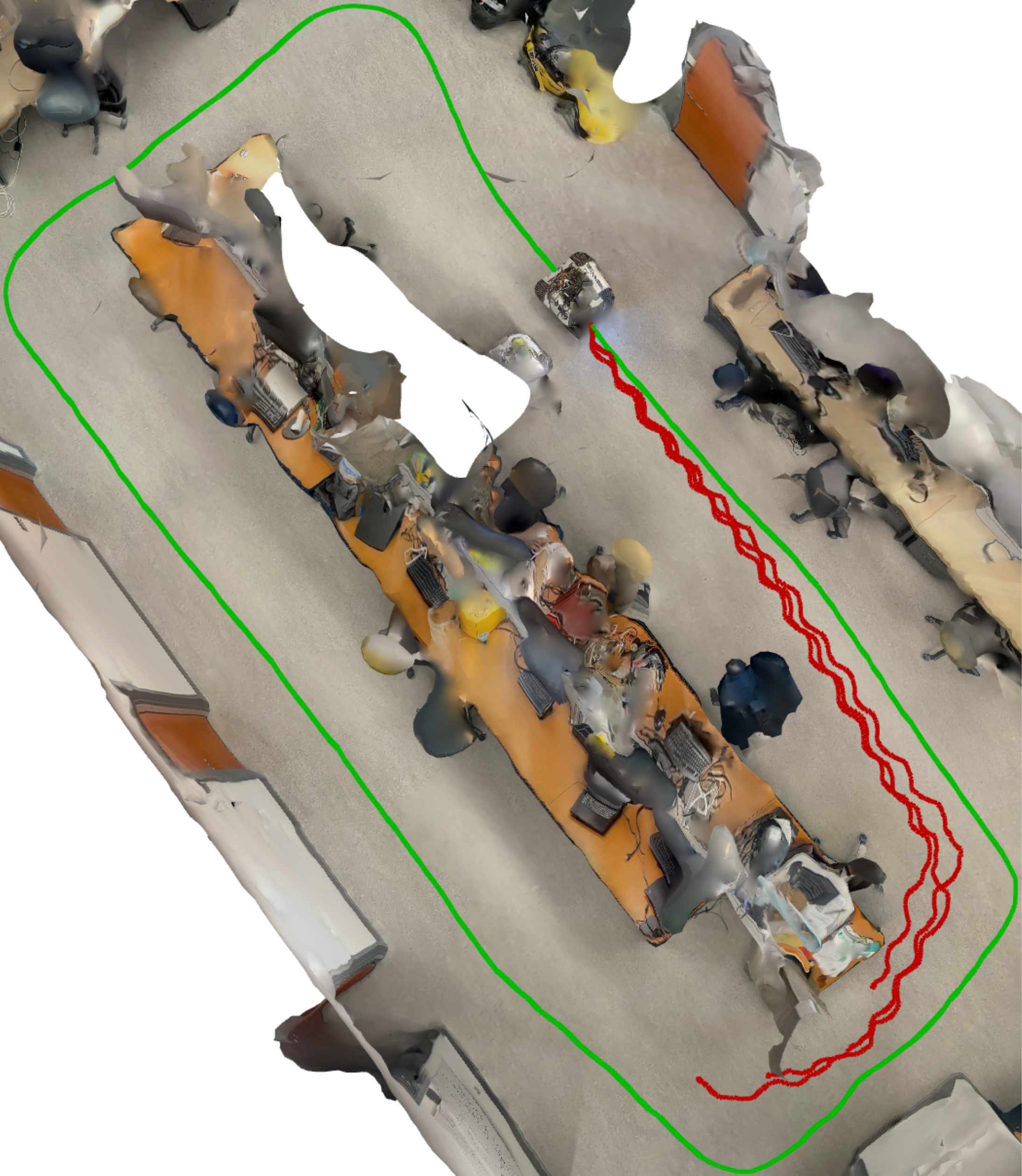} \\
        \end{tabular}
    \caption{Last three \textbf{Office loop} trajectories for \prettygnm{}, \prettyvint{}, \prettynomad{} and \prettybridger{} (see Table~\ref{tab:precision_metrics_all_envs},~\ref{tab:generalization_metrics_all_env} and~\ref{tab:collision}) with the \prettyref{}.}
    \label{fig:loop_mesh}
\end{figure}

\textbf{Office Loop:} A full loop of an office environment (\textit{27.960m, Bunker}) with chairs, desks, computers, and drawers, requiring long-horizon navigation in a cluttered indoor setting (\textit{see Figure~\ref{fig:loop_mesh}}).

\textbf{Arena:} A doorway navigation task (\textit{20.051m, Spot}) requiring the robot to exit one door and enter the third door on the left. Two additional open doors precede the goal, testing visual discrimination; the target door is distinguished by the highest number of distinctive visual features (\textit{see Figure~\ref{fig:arena_mesh}}).

\textbf{Snow:} An outdoor snowy parking lot (\textit{18.964m, Bunker}) representing a significant out-of-distribution shift, evaluating model robustness and generalization under adverse environmental conditions (\textit{see Figure~\ref{fig:snow_mesh}}).

\subsection{Implementation details}


All models were deployed on Bunker and Spot (Nvidia AGX Xavier/Orin) using the open-source pretrained weights exported to ONNX, enabling optimized onboard inference. Data was collected via ROS/ROS~2 as bag files across all environments (see \Cref{fig:meet_the_envs}).

For each environment, a reference trajectory was first recorded by manually piloting the robot, capturing RGB images from an onboard fisheye camera and poses from SLAM-LVI-SAM (LiDAR+IMU) under ROS and SuperOdometry (LiDAR-only) under ROS~2~\cite{lvisam21, zhao2021super} (see \Cref{fig:pitch}). The images are sampled at fixed intervals along this trajectory to form a linear topological map.
The models were then tasked with autonomously following this map, with the final node of the created topomap as the goal. 
Note that Bunker lacks a collision avoidance mechanism. Spot's in-built controller prevents collisions, but the model's failures to reach the goal indicate overshoot or the robot becoming stuck.

\subsection{Dataset Composition}

The dataset is composed of trajectories across the five presented environments where each deployment trial logged odometry, predicted distances, node predictions, goal detection, CPU/GPU/memory usage, and inference time.

\subsection{Visual perturbations}

Corridor has two variants, which introduce motion blur and sunflare perturbations~\cite{albumentations} to the topological map, replicating fast motion and adverse lighting conditions, while RGB observations remain unmodified, eliminating real-time image processing overhead. We only evaluate models that succeed in the standard Corridor, as assessing visual perturbations for models that already failed the standard version is uninformative.

\section{Results}
\begin{figure*}[t]
    \centering
    \small 
    \begin{tabular}{@{}p{0.10\linewidth}*{5}{p{0.10\linewidth}}@{}}
        \centering\textbf{Goal Image} & \centering\prettygnm{} & \centering\prettyvint{} & \centering\prettynomad{} & \centering\arraybackslash\prettybridger{} & \centering\arraybackslash\prettycross{}\\
        \midrule
        
        \centering\includegraphics[width=\linewidth]{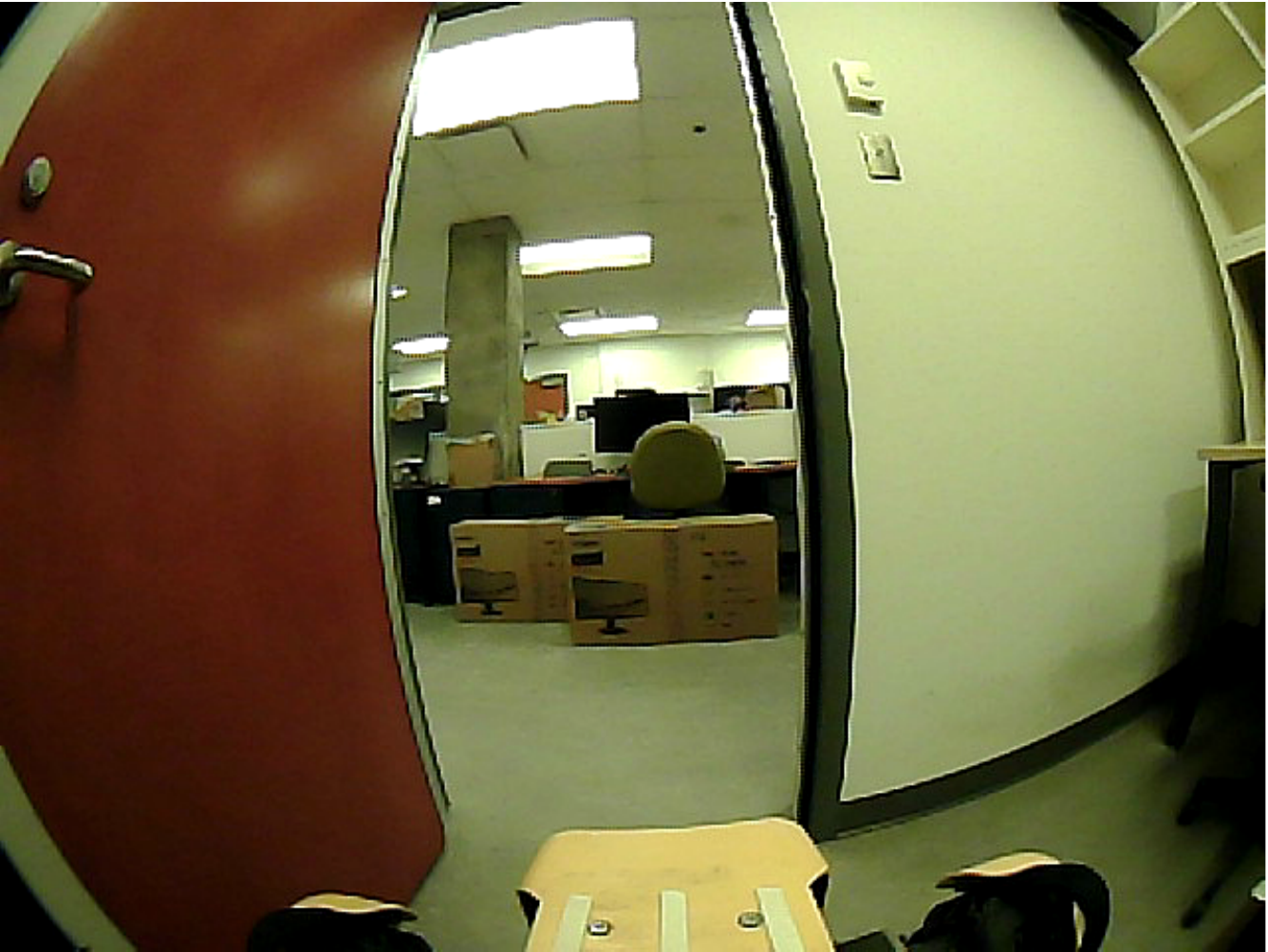}
        & \centering\includegraphics[width=\linewidth]{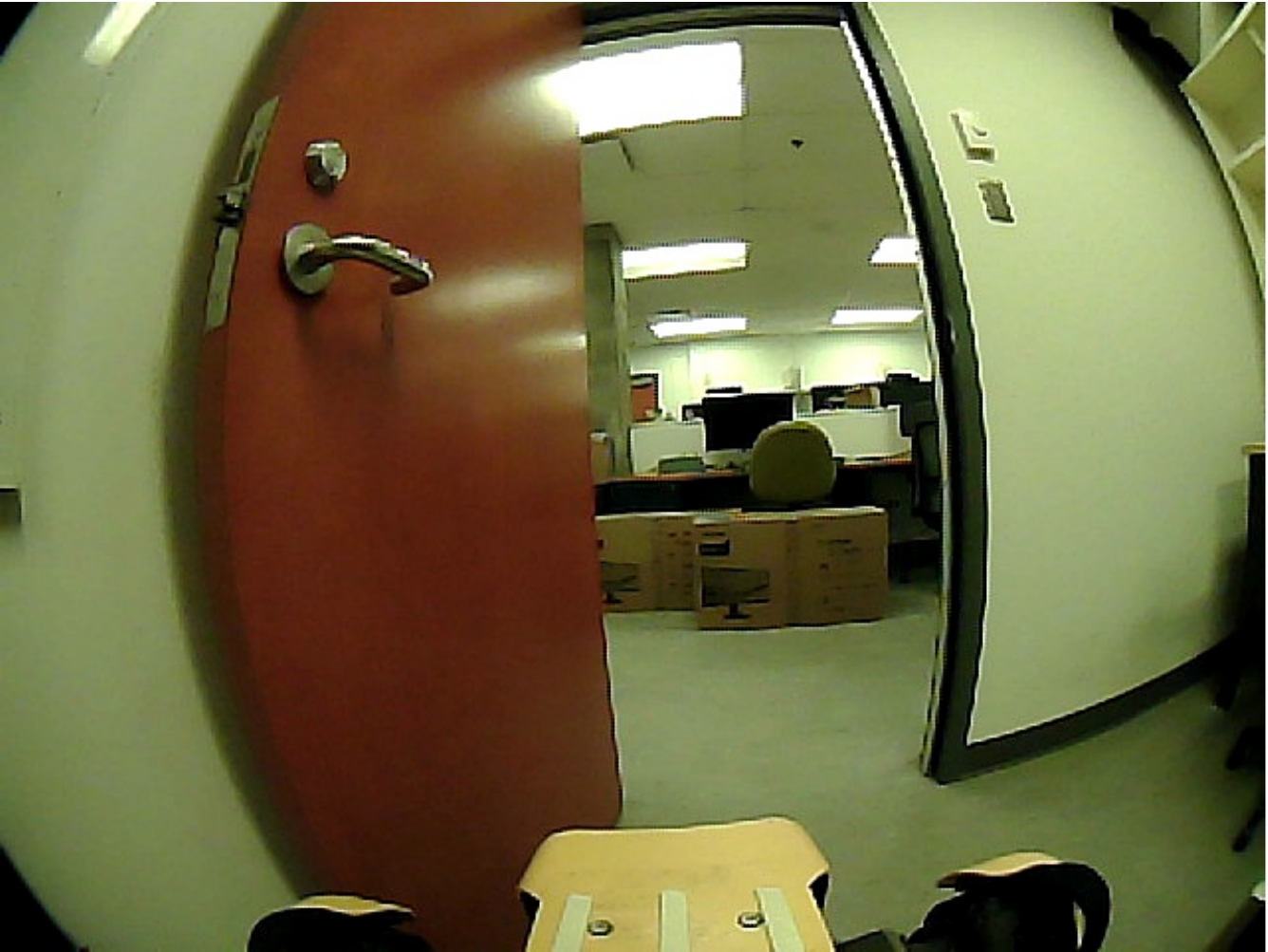} 
        & \centering\includegraphics[width=\linewidth]{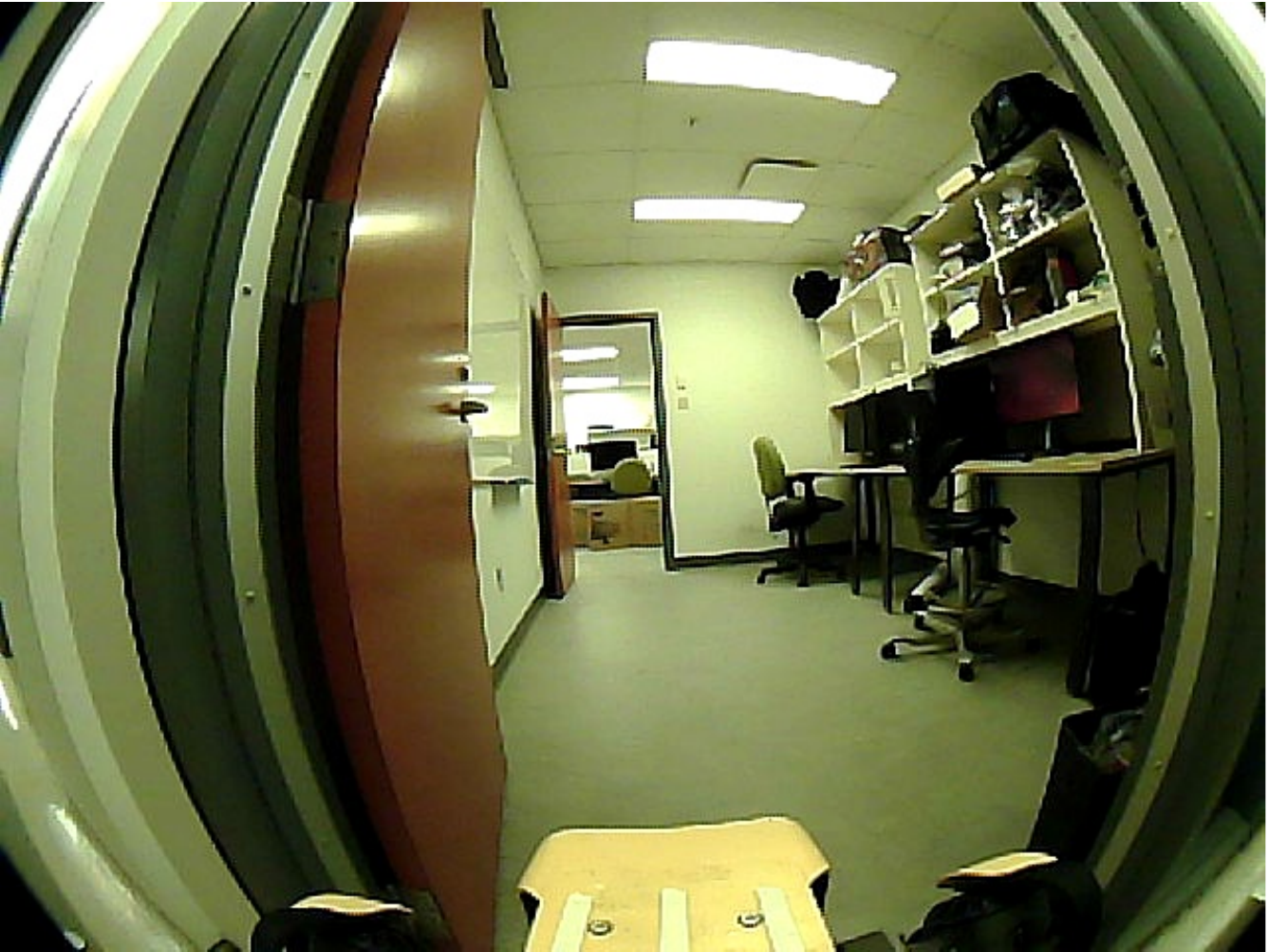} 
        & \centering\includegraphics[width=\linewidth]{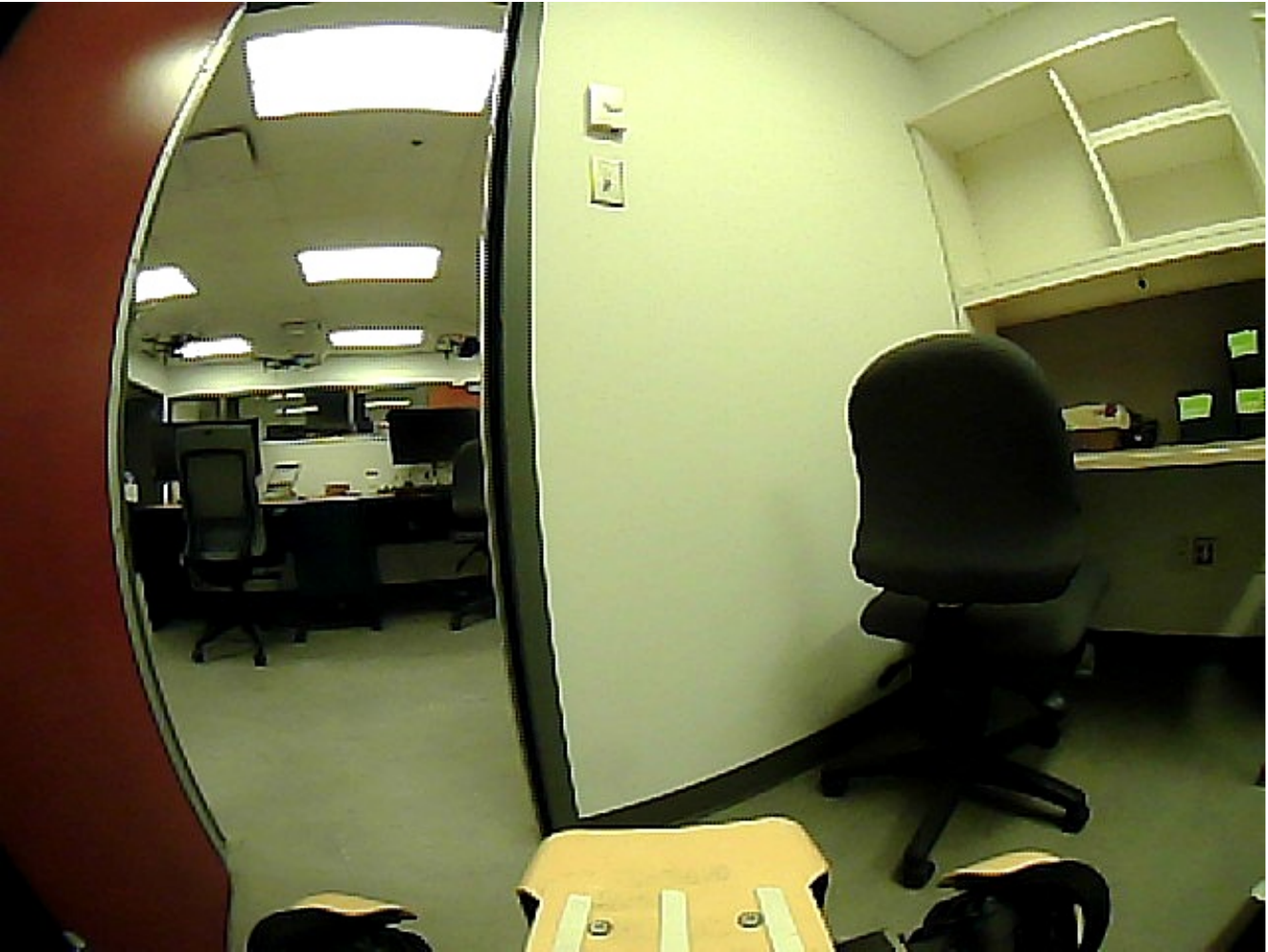} 
        & \centering\arraybackslash\includegraphics[width=\linewidth]{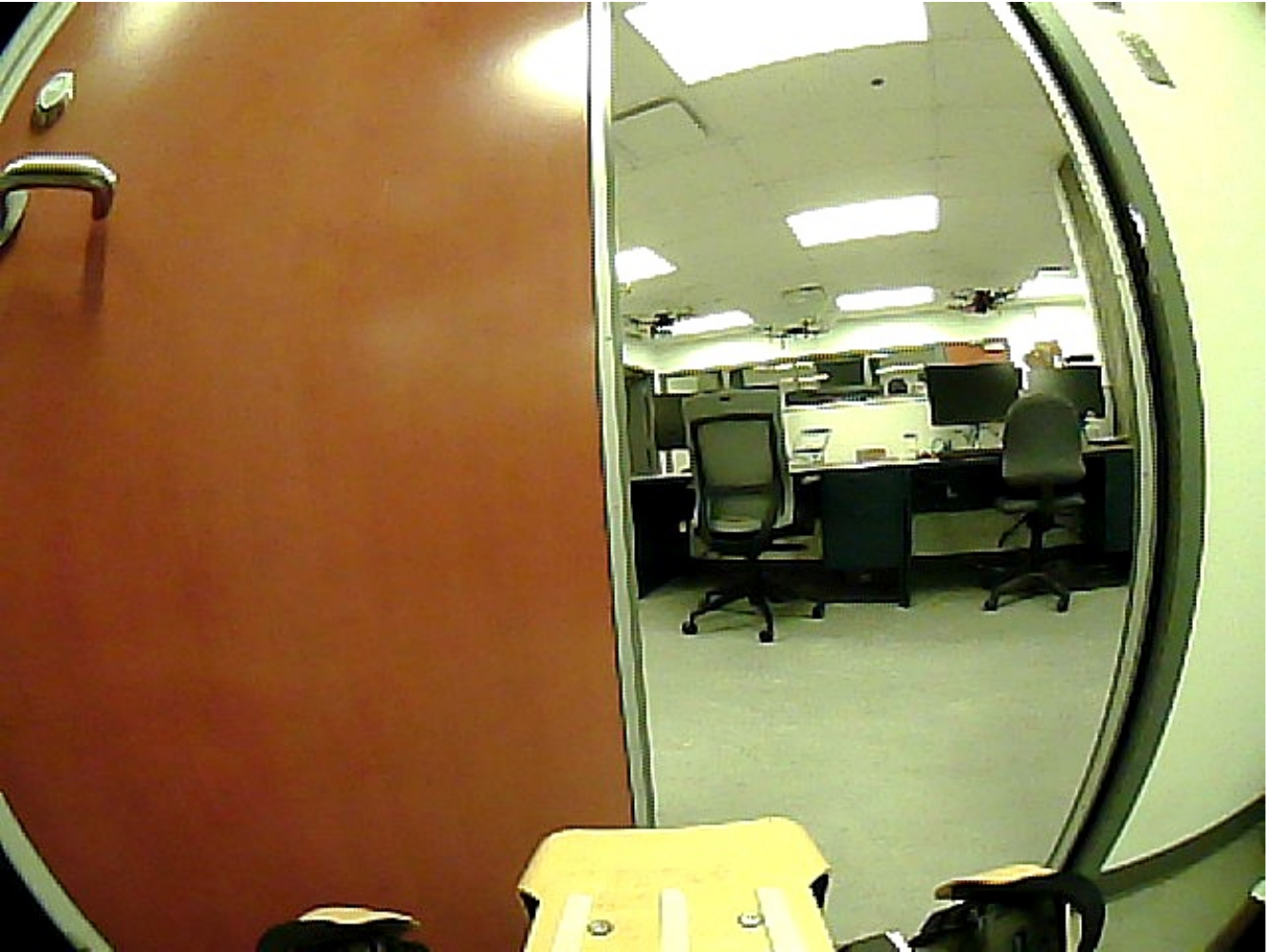} 
        &
        \centering\arraybackslash\includegraphics[width=\linewidth]{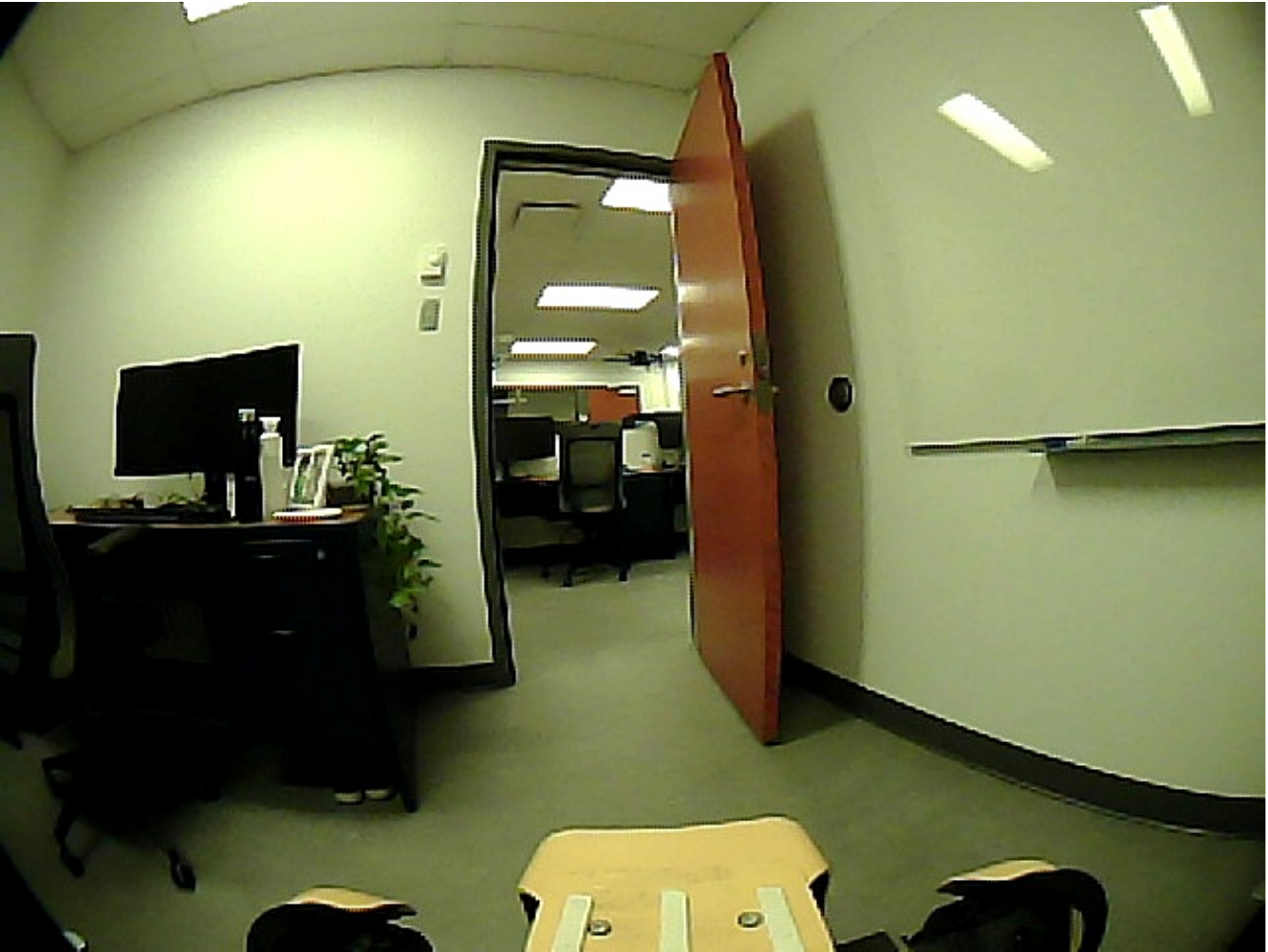} \\
        \midrule
        
        LPIPS $\downarrow$ \newline
        DSSIM $\downarrow$ \newline
        PSNR $\uparrow$
        & \centering
         0.57 $\pm$ 0.04 \newline 
         0.08 $\pm$ 0.01 \newline 
         11.84 $\pm$ 0.58 
        & \centering
        0.67 $\pm$ 0.00 \newline 
        0.29 $\pm$ 0.12 \newline 
        9.34 $\pm$  0.10 
        & \centering
        0.68 $\pm$ 0.06 \newline 
        0.27 $\pm$ 0.10 \newline 
        9.51 $\pm$ 1.38 
        & \centering
        0.64 $\pm$  0.08 \newline 
        0.27 $\pm$  0.12 \newline 
        10.19 $\pm$  1.03 
        & \centering\arraybackslash
        0.67 $\pm$  0.03 \newline 
        0.28 $\pm$  0.03 \newline 
        10.35 $\pm$  0.65 \\ 
        \bottomrule
    \end{tabular} 
    
    \begin{tabular}{@{}p{0.10\linewidth}*{5}{p{0.10\linewidth}}@{}}
        \centering\textbf{Goal Image} & \centering\prettygnm{} & \centering\prettyvint{} & \centering\prettynomad{} & \centering\arraybackslash\prettybridger{} \\
        \midrule
        
        \centering\includegraphics[width=\linewidth]{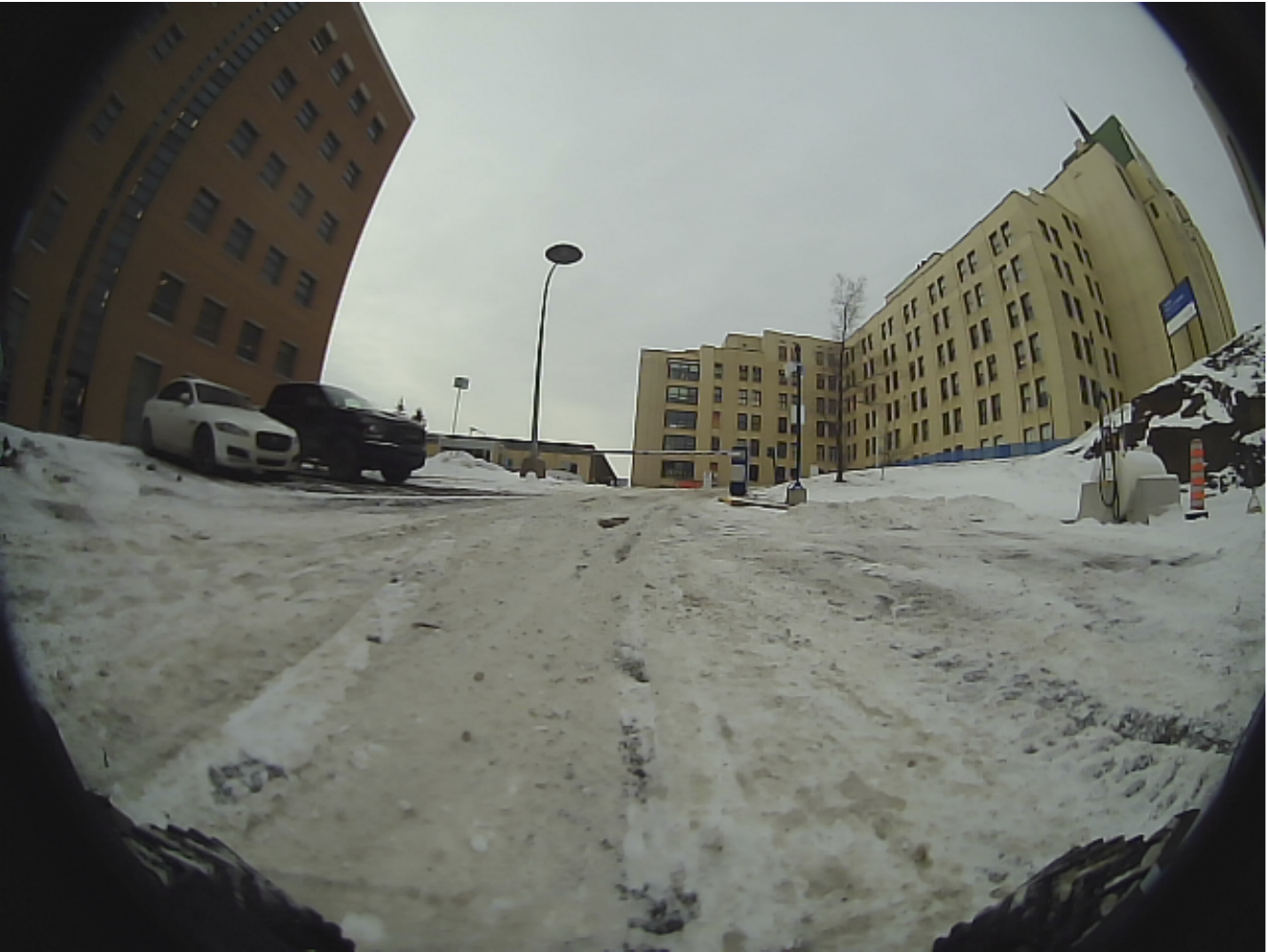}
        & \centering\includegraphics[width=\linewidth]{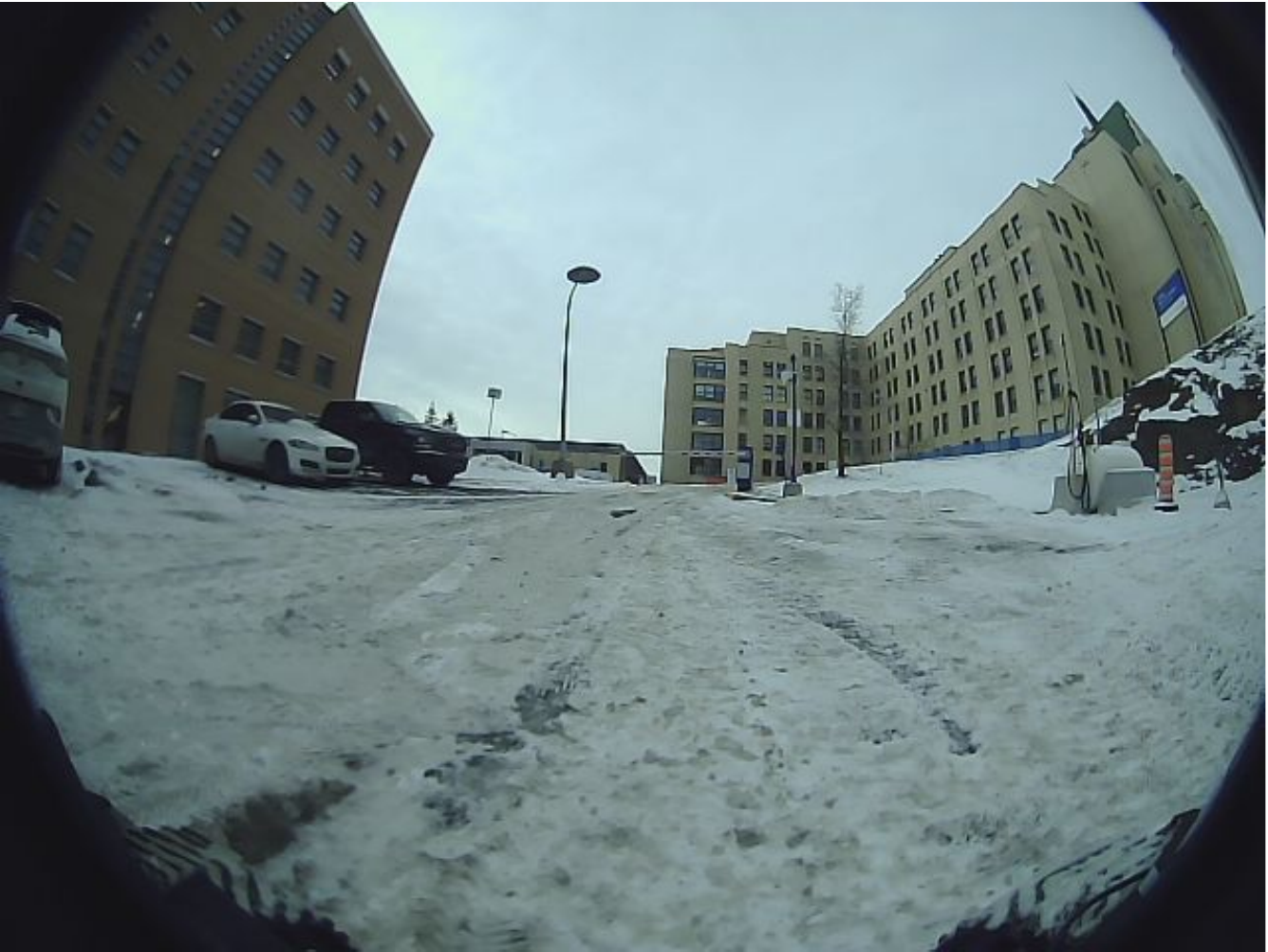} 
        & \centering\includegraphics[width=\linewidth]{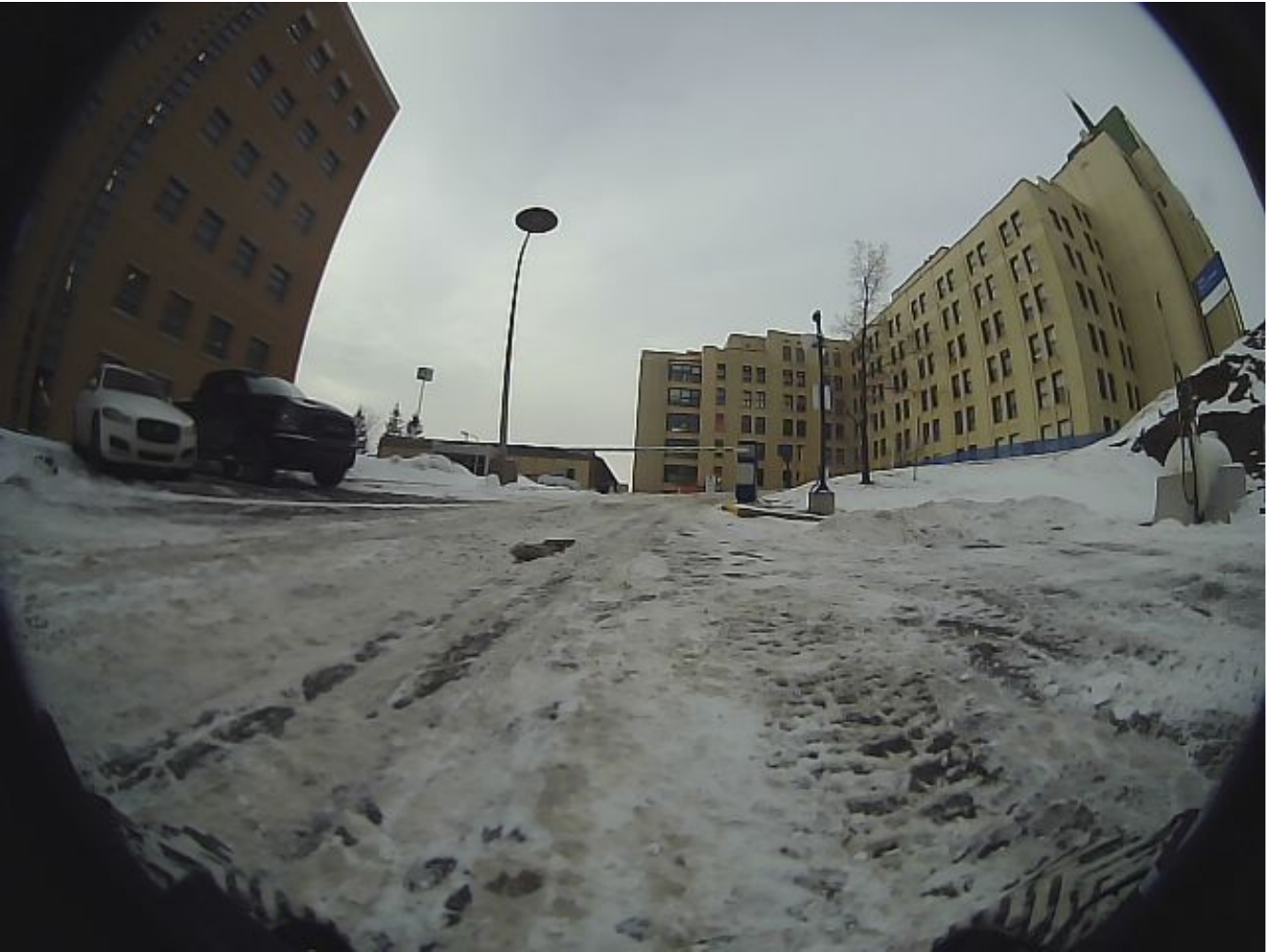} 
        & \centering\includegraphics[width=\linewidth]{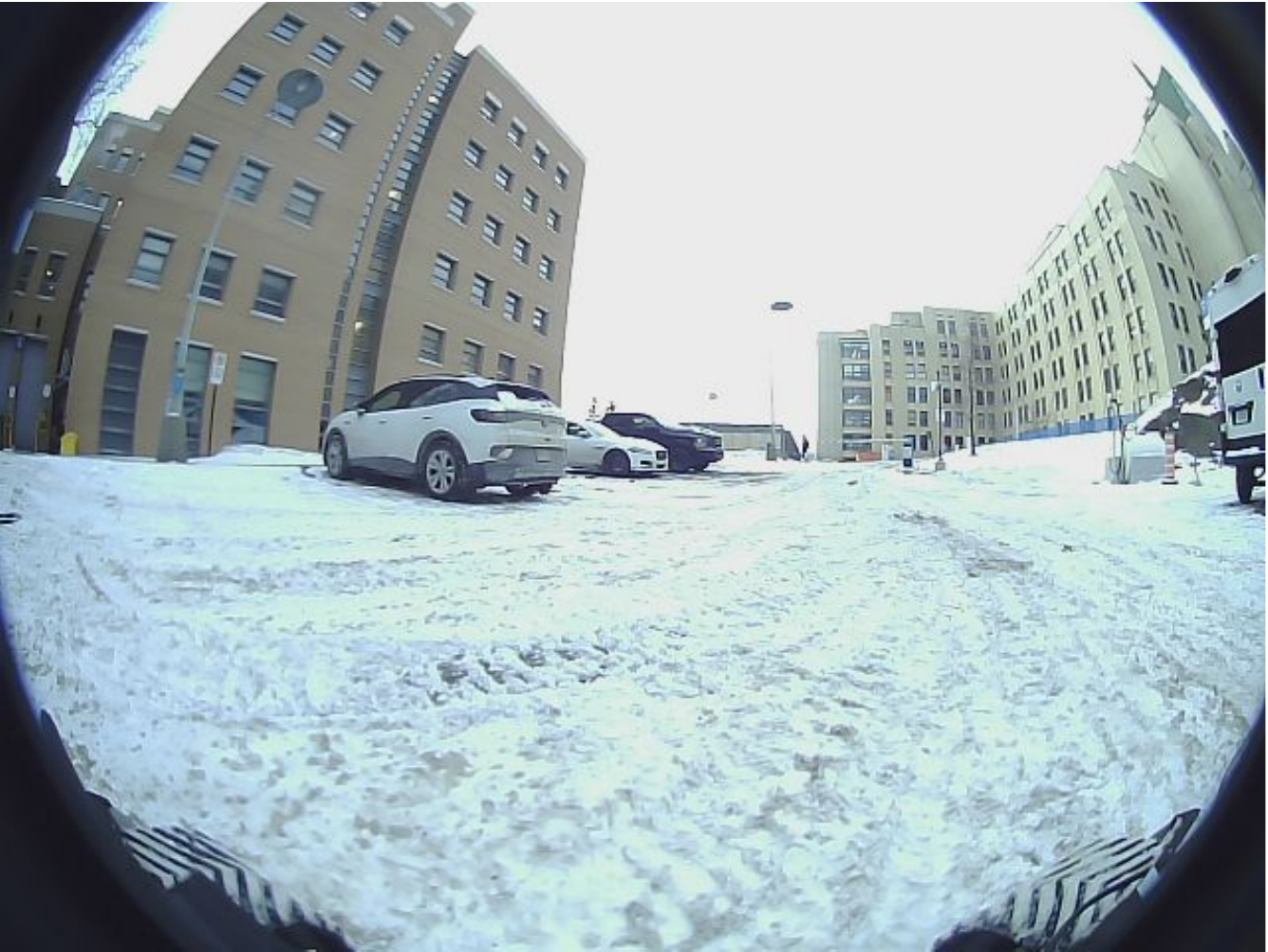} 
        & \centering\arraybackslash\includegraphics[width=\linewidth]{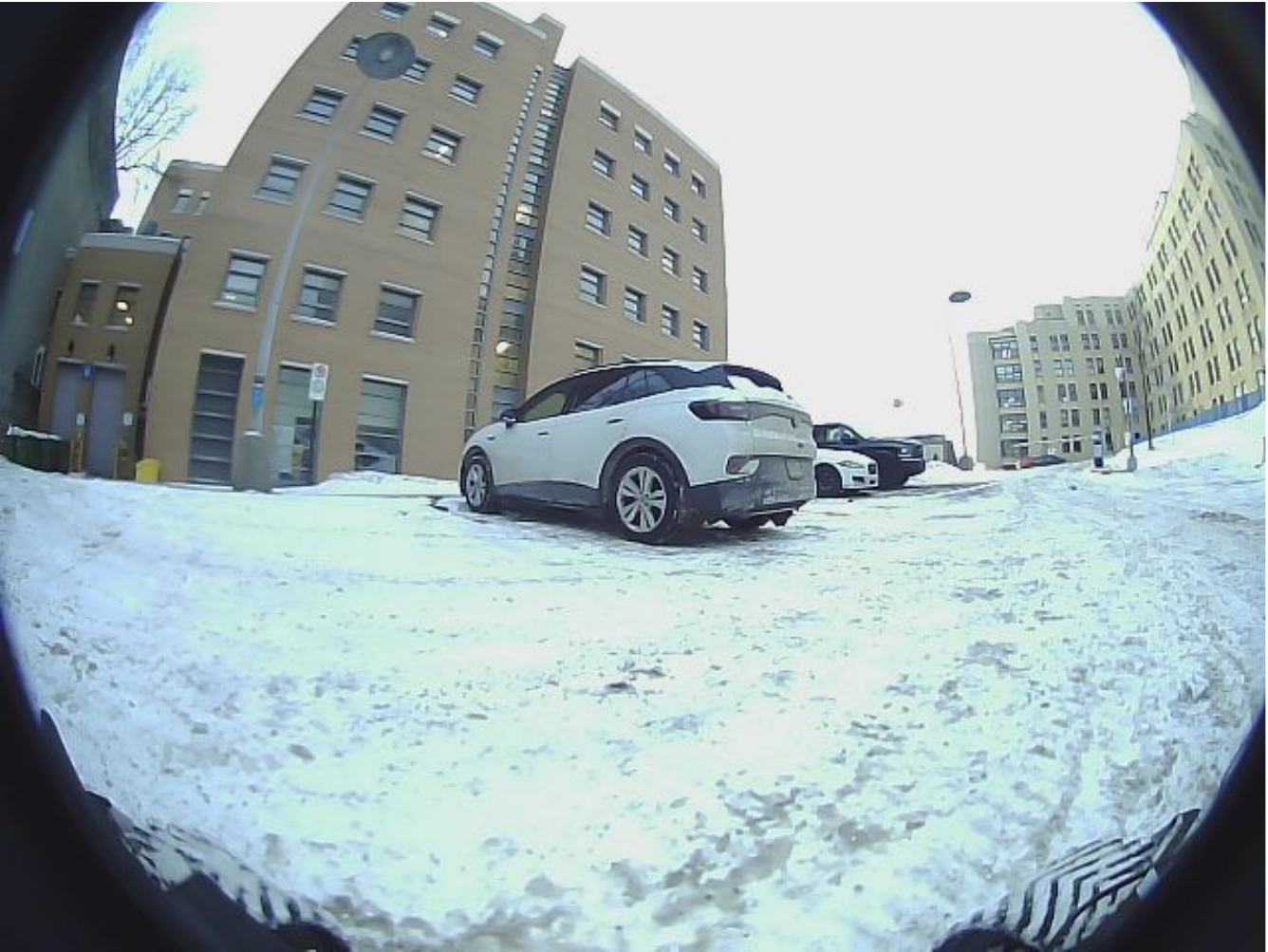} \\
        \midrule
        
        LPIPS $\downarrow$ \newline
        DSSIM $\downarrow$ \newline
        PSNR $\uparrow$
        & \centering\small 
        0.36 $\pm$ 0.02 \newline 
        0.09 $\pm$  0.00 \newline 
        16.08 $\pm$ 1.26 
        & \centering\small 
        0.44 $\pm$ 0.09 \newline 
        0.13 $\pm$ 0.07 \newline 
        15.85 $\pm$ 1.57 
        & \centering\small 
        0.53 $\pm$ 0.05 \newline 
        0.21 $\pm$ 0.03 \newline 
        11.84 $\pm$  3.54 
        & \centering\arraybackslash\small 
        0.48 $\pm$ 0.12 \newline 
        0.20 $\pm$  0.11 \newline 
        13.38 $\pm$  3.77 \\ 
        \bottomrule
    \end{tabular}
    
    \caption{Image quality metrics (LPIPS, PSNR, DSSIM) for goal-predicted cases in the \textbf{Arena} (Top) and \textbf{Snow} (Bottom) environments.}
    \label{fig:img_metrics_all}
\end{figure*}

\begin{table*}[t]
    \centering
    \caption{Topological Node Error (\textit{n.err.}) and Predicted Goal Counter (\textit{g.pred.}) across environments and models.}
    \begin{tabular}{l*{2}{c}*{2}{c}*{2}{c}*{2}{c}*{2}{c}}
    \toprule
    & \multicolumn{2}{c}{\textbf{Corridor}} 
    & \multicolumn{2}{c}{\textbf{Office loop}} 
    & \multicolumn{2}{c}{\textbf{Arena}} 
    & \multicolumn{2}{c}{\textbf{Stairs}} 
    & \multicolumn{2}{c}{\textbf{Snow}} \\
    \cmidrule(lr){2-3} \cmidrule(lr){4-5} \cmidrule(lr){6-7} \cmidrule(lr){8-9} \cmidrule(lr){10-11}
    \textbf{Method} & n.err. & g.pred. & n.err. & g.pred. & n.err. & g.pred. & n.err. & g.pred. & n.err. & g.pred. \\
    \midrule
    \rowcolor{gray!15}
    \prettygnm{} & 0.36$\pm$0.49 & 3/3 & 3.23$\pm$5.47 & 5/10 & 6.94$\pm$1.96 & 2/3 & 0.89$\pm$0.55 & 3/3 & 1.95$\pm$1.05 & 3/3 \\
    \rowcolor{gray!5}
    \prettyvint{} & 0.43$\pm$0.49 & 3/3 & 7.34$\pm$13.58 & 9/10 & 7.43$\pm$3.05 & 2/5 & 0.94$\pm$0.66 & 3/3 & 3.82$\pm$5.52 & 3/3 \\
    \rowcolor{orange!15}
    \prettynomad{} & 0.49$\pm$0.70 & 0/3 & 8.53$\pm$15.04 & 8/10 & 10.59$\pm$7.76 & 5/5 & 0.98$\pm$0.60 & 3/3 & 3.95$\pm$3.65 & 3/3 \\
    \rowcolor{orange!5}
    \prettybridger{}  & 0.72$\pm$0.44 & 1/3 & 2.37$\pm$2.59 & 1/10 & 11.04$\pm$6.91 & 3/5 & 2.01$\pm$1.29 & 3/3 & 2.47$\pm$1.92 & 3/3 \\
    \prettycross{}  & 0.41$\pm$0.49 & 0/3 & - & - & 13.21$\pm$6.97 & 5/5 & - & - & - & - \\
    \bottomrule
    \end{tabular}    \label{tab:generalization_metrics_all_env}
\end{table*}

\begin{table}[b]
    \centering
    \caption{Collision (\textit{col.}) results for Corridor and Office loop.}
    \begin{tabular}{lcc}
    \toprule
    \textbf{Method} & \textbf{Corridor} & \textbf{Office loop} \\
    \cmidrule(lr){2-2} \cmidrule(lr){3-3}
    & col. & col. \\
    \midrule
    \rowcolor{gray!15}
    \prettygnm{} & 0/3 & 5/10 \\
    \rowcolor{gray!5}
    \prettyvint{} & 0/3 & 1/10 \\
    \rowcolor{orange!15}
    \prettynomad{} & 3/3 & 2/10 \\
    \rowcolor{orange!5}
    \prettybridger{}  & 2/3 & 9/10 \\
    \prettycross{}  & 3/3 & - \\
    \bottomrule
    \end{tabular}
    \label{tab:collision}
\end{table}

\begin{figure*}[h]
    \centering
  \includegraphics[width=0.65\textwidth]{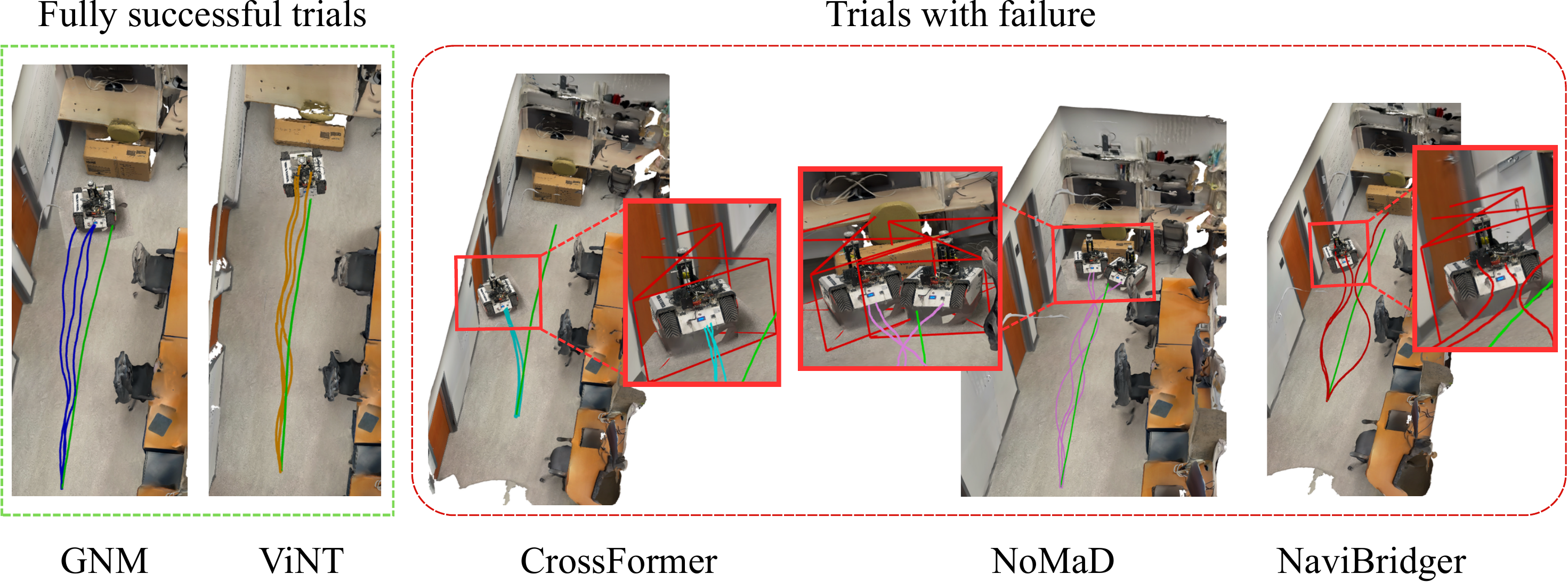}
  \caption{\textbf{Corridor deployment results.} Green (left): goal reached without collision in all trials. Red (right): at least one collision occurred. See Table~\ref{tab:collision} for details. (Note: overlapping collisions at the same location are shown only once for clarity).}
  \label{fig:corridor_figure}
\end{figure*}

\subsection{In-Distribution Performance Analysis}
Table~\ref{tab:precision_metrics_all_envs} reports distance to goal and path length across models and environments, while Table~\ref{tab:generalization_metrics_all_env} covers topological node error and goal prediction. Collisions occurrence are reported in Table~\ref{tab:collision}. Perception metrics are shown in Figure~\ref{fig:img_metrics_all}, and visual perturbation results are in Table~\ref{tab:perturbation_easy_office}.
The indoor environments include Corridor, Office loop, Arena, Stairs, while Snow is an outdoor environment.

\subsubsection{Collision Hotspots: How Lack Of Geometric Understanding Impacts Navigation Performance} 

Even when operating in a familiar office setting, all models show significant limitations. In the corridor environment, CrossFormer, NoMaD, and, NaviBridger fail at both goal prediction (see Table~\ref{tab:generalization_metrics_all_env}) and collision avoidance (see Table~\ref{tab:collision} and Figure~\ref{fig:corridor_figure}). In contrast, ViNT and GNM succeed in all 3 trials, reaching close to the goal position on average. While models consistently move along the reference trajectory (4-5 m average path length), the key dissociation is clear: models can follow paths but fail to recognize goals images and avoid collisions, even in conditions similar to the training environments.


These collisions reveal a fundamental lack of geometric understanding: across all architectures (feedforward, transformer and diffusion-based), vision-based reasoning alone cannot reliably encode obstacle geometry. Failures worsen in the complex office loop setting (see Table~\ref{tab:collision}), where models collide with door frames, desks, chairs, and pillars. Two failure modes emerge: insufficient collision examples in training data, and the absence of explicit geometric reasoning. 

Subsequent experiments offload collision avoidance to a low-level planner on the Spot robot, isolating visual navigation performance from geometric failures.

\subsubsection{Precision vs. Completion: Analyzing Prediction Patterns}

In the stairs environment (see Table~\ref{tab:precision_metrics_all_envs}), GNM, ViNT and NoMaD closely follow the reference trajectory and consistently identify the goal image. Notably, NoMaD shows improved goal recognition compared to the corridor setting, achieving 3/3 successful goal recognition with substantially lower distance-to-goal. This prompts a critical question: what drives model failures? To this end, we examine a setting with semantically similar images with subtle feature variations (the \textit{Arena} environment).

\begin{figure}[htbp]
    \centering
       \begin{tabular}{@{}c@{\hspace{0.5em}}c@{}}
            \includegraphics[width=0.39\columnwidth]{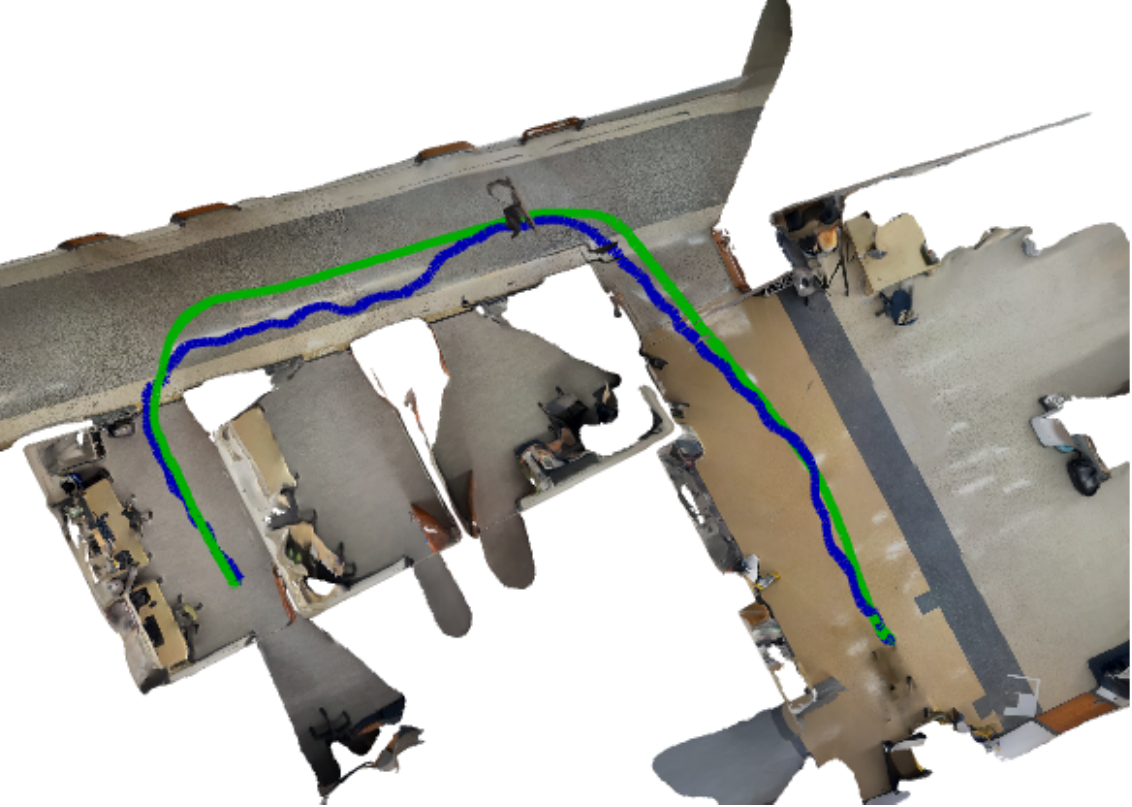} &
            \includegraphics[width=0.39\columnwidth]{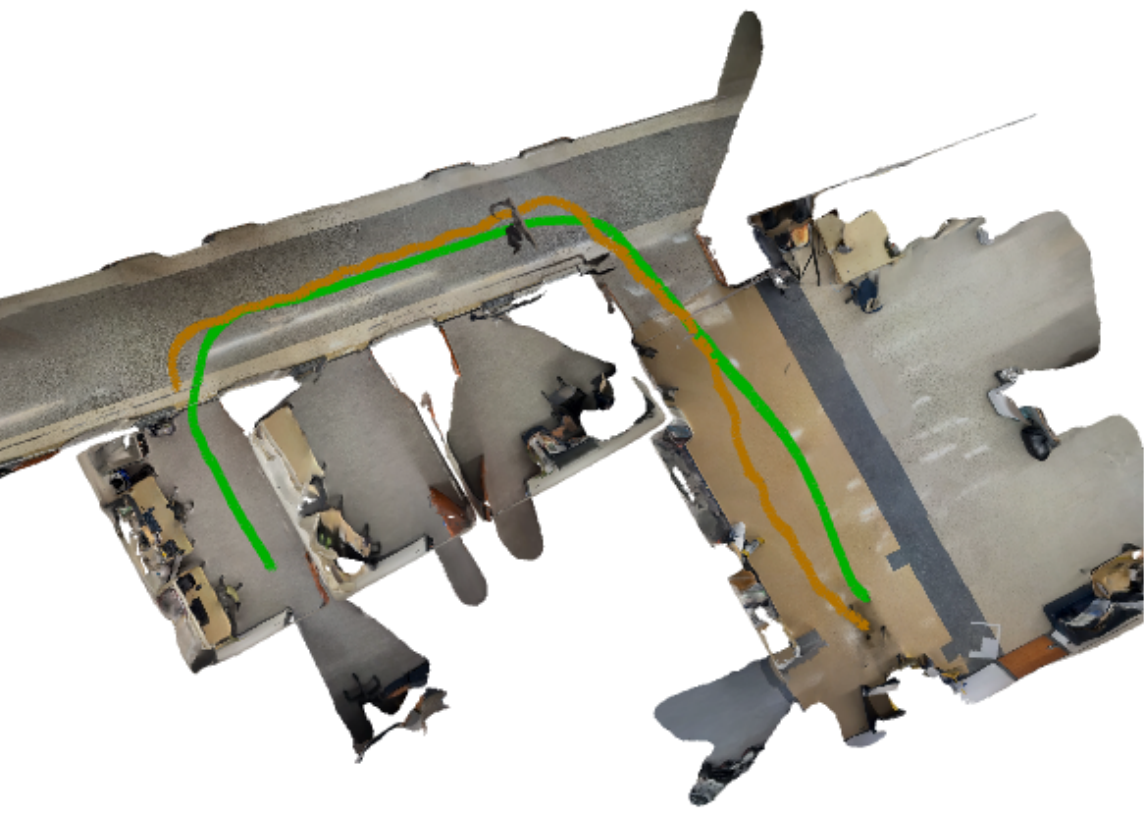} \\
            \includegraphics[width=0.39\columnwidth]{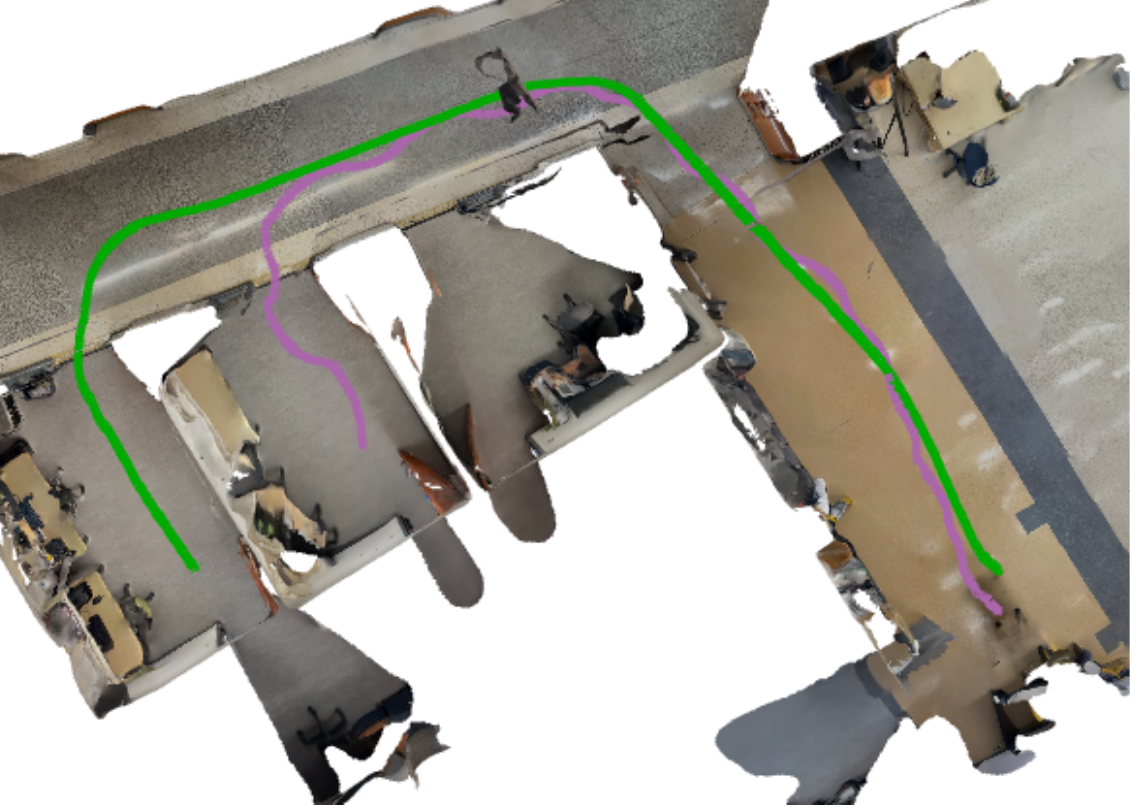} &
            \includegraphics[width=0.39\columnwidth]{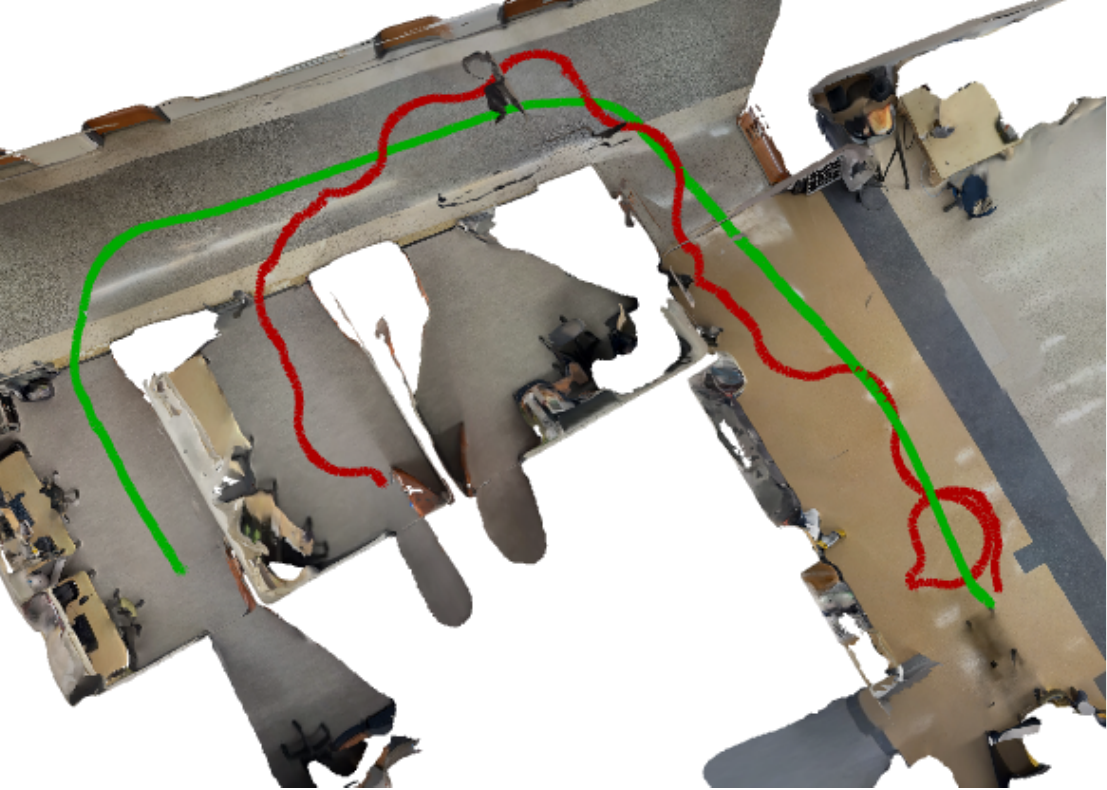} \\
            \multicolumn{2}{c}{\includegraphics[width=0.37\columnwidth]{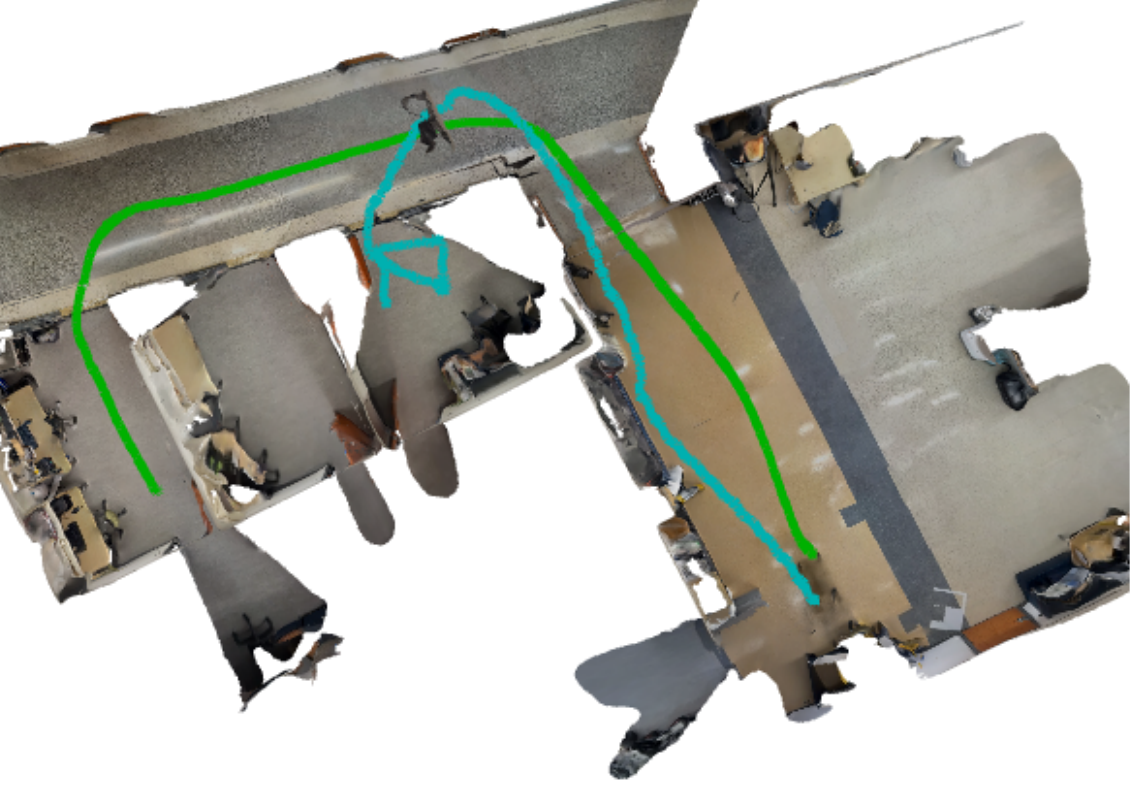}} \
        \end{tabular}
    \caption{One \textbf{Arena} trajectory for \prettygnm{}, \prettyvint{}, \prettynomad{}, \prettybridger{}, and \prettycross{} (see Table~\ref{tab:precision_metrics_all_envs} and~\ref{tab:generalization_metrics_all_env}) with the \prettyref{}.}
    \label{fig:arena_mesh}
\end{figure}

In the Arena (see Figure~\ref{fig:arena_mesh}) environment, NoMaD and ViNT achieve the highest path lengths, demonstrating strong trajectory imitation, yet both exhibit consistent failure to attain the goal, reflected in higher average distances to goal (see Table~\ref{tab:generalization_metrics_all_env}). Their shared EfficientNet-B0 encoder likely contributes to this common failure. 

Similarly, CrossFormer and NaviBridger show severe failure to reach the goal, with an average goal distances of 10 meters and path lengths exceeding near 30 meters against a 20 meters reference trajectory. This suggests that models struggle to quantify progress through the topological map. In contrast, GNM's simpler MobileNetV2 encoder achieves more consistent goal prediction. These results reveal \textit{what} fails: failure to reach the goal, premature goal prediction, trajectory deviation, but not \textit{why}. We hypothesize that performance gap between architecturally similar models (NoMaD vs. ViNT) and GNM's surprising robustness suggests encoding quality matters more than architectural complexity.


\begin{figure}[htbp]
    \centering
       \begin{tabular}{@{}c@{\hspace{0.5em}}c@{}}
            \includegraphics[width=0.41\columnwidth]{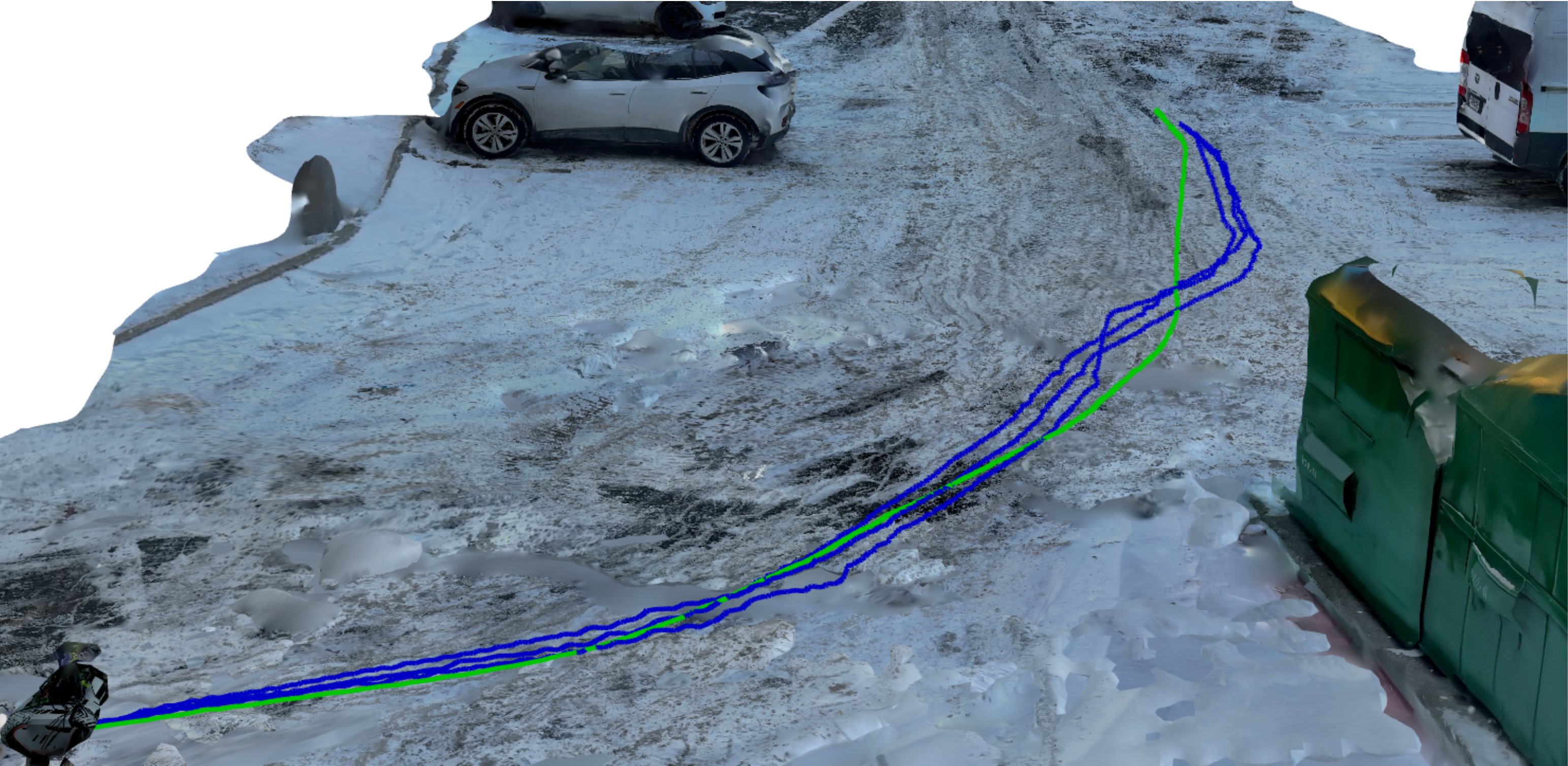} &
            \includegraphics[width=0.41\columnwidth]{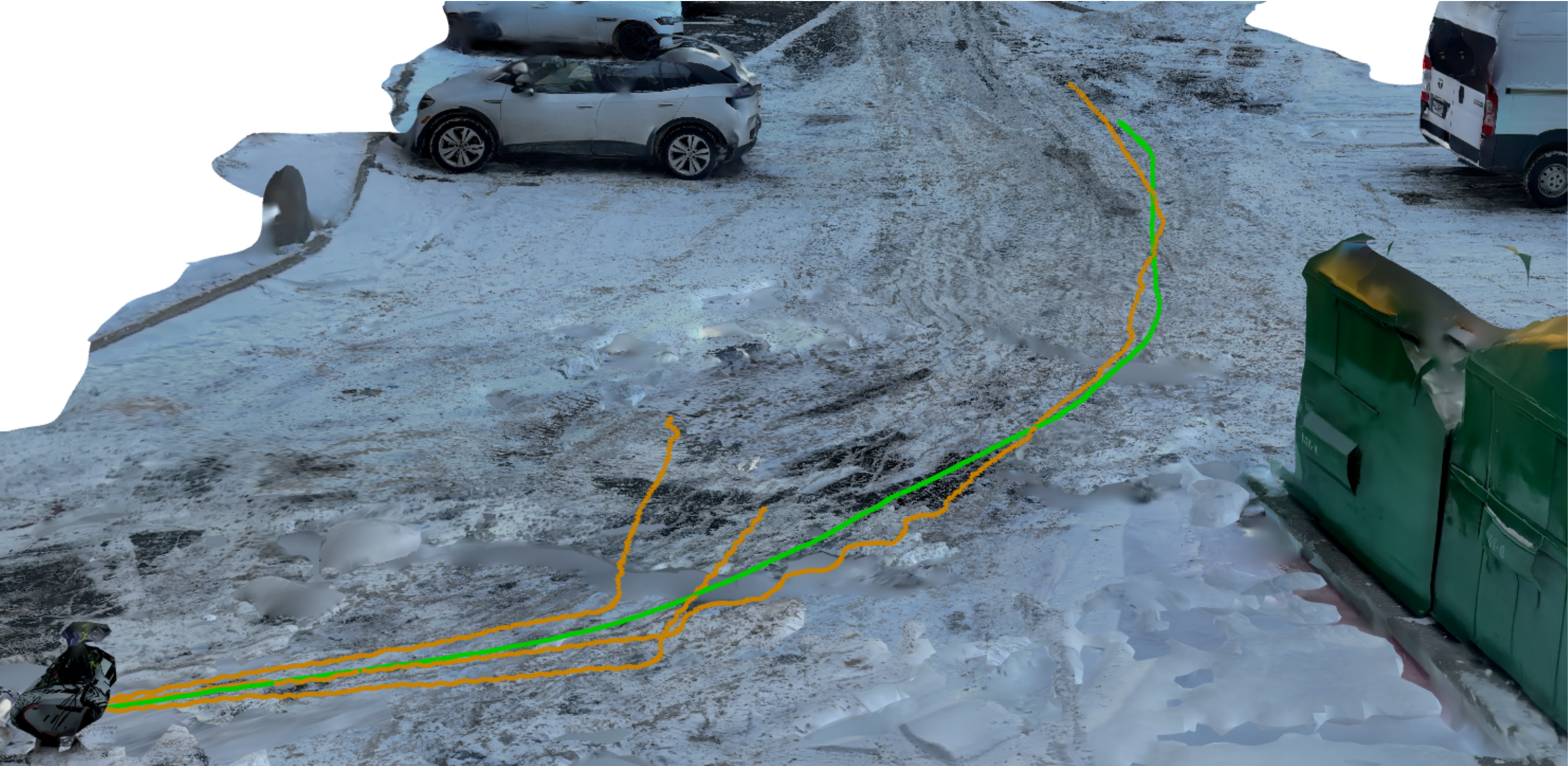} \\
            \includegraphics[width=0.41\columnwidth]{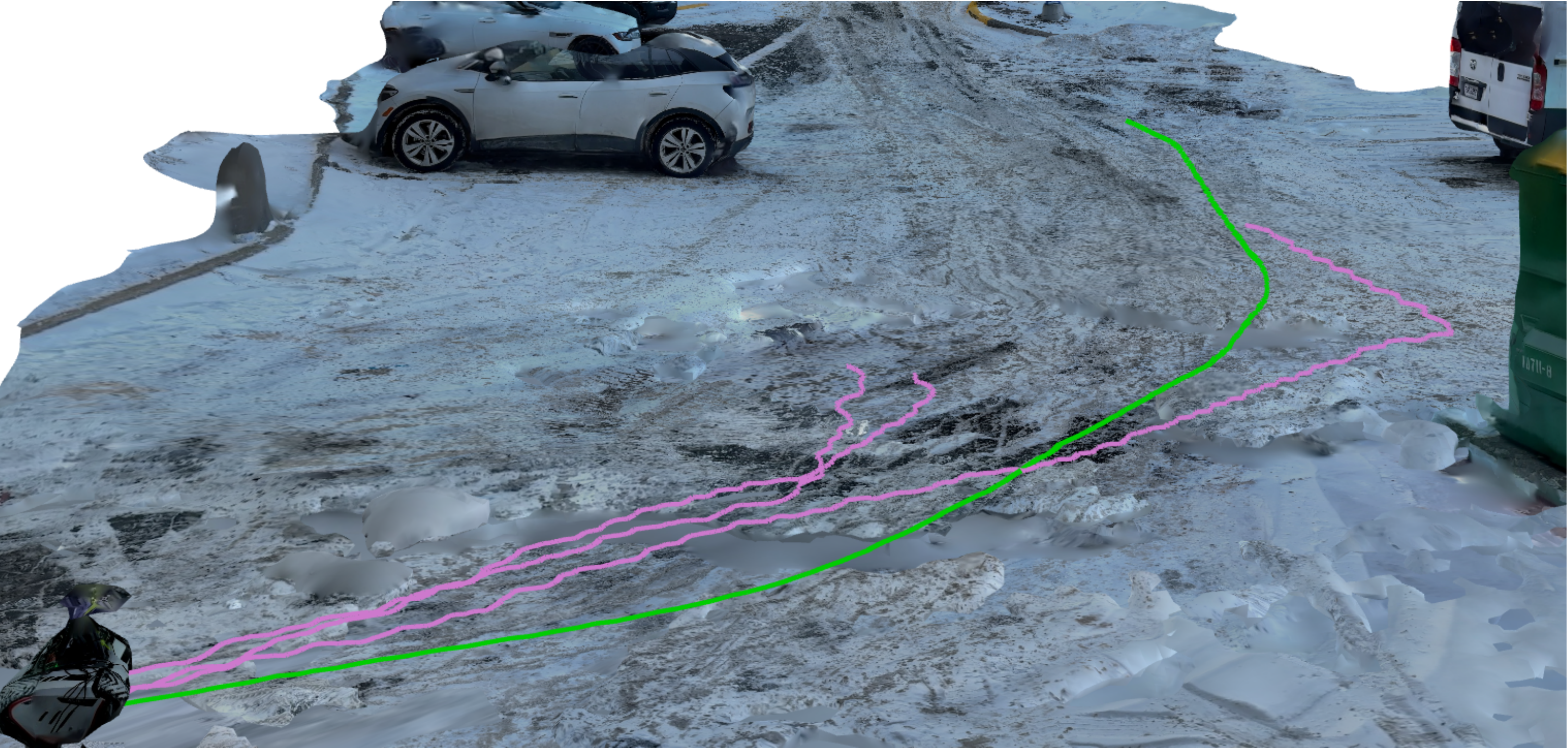} &
            \includegraphics[width=0.41\columnwidth]{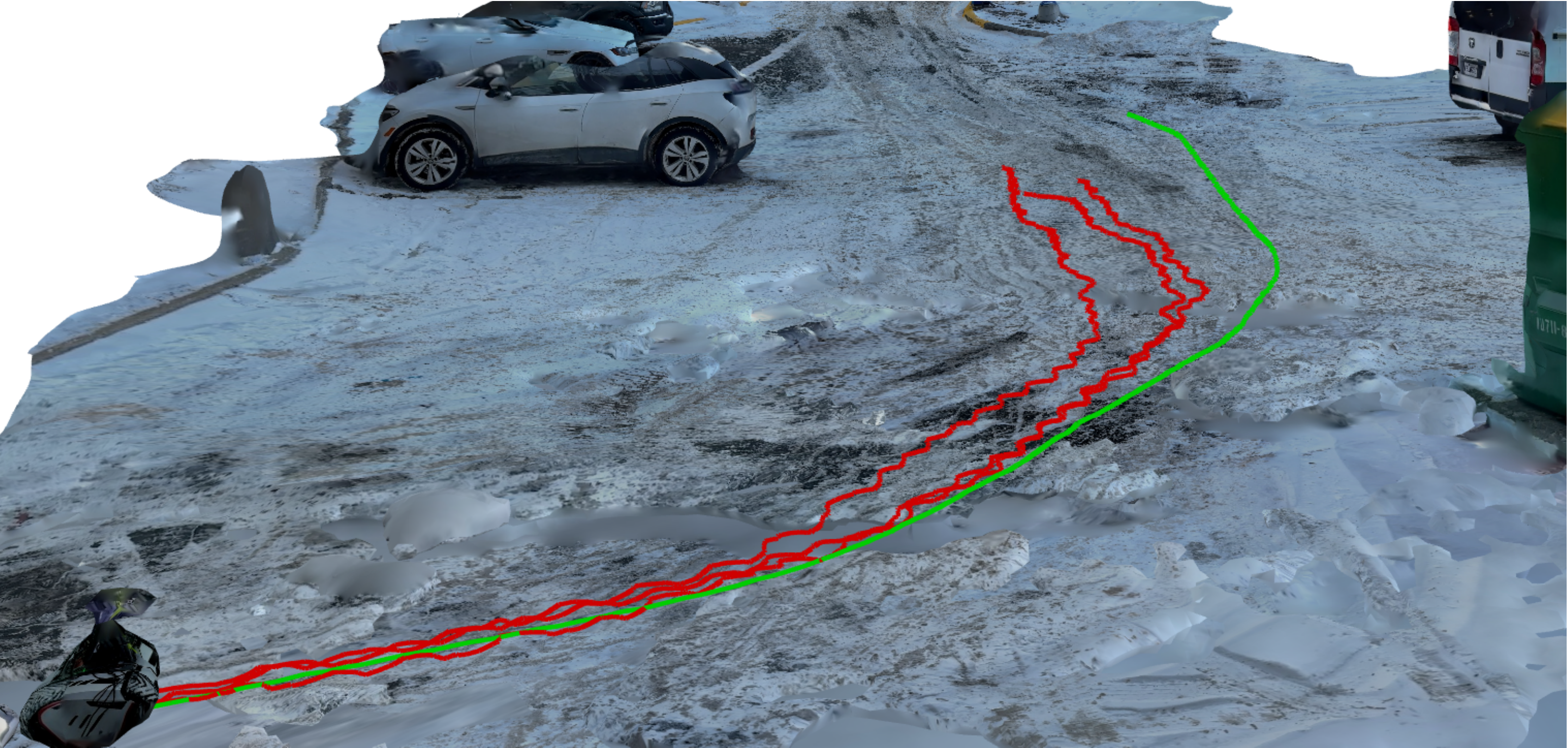} \\
        \end{tabular}
    \caption{All \textbf{Snow} trajectories for \prettygnm{}, \prettyvint{}, \prettynomad{} and \prettybridger{} (see Table~\ref{tab:precision_metrics_all_envs} and~\ref{tab:generalization_metrics_all_env}) with the \prettyref{}.}
    \label{fig:snow_mesh}
\end{figure}

\subsection{From Pixels to Waypoints: The Role of Visual Encoding}

We now consider how the vision encoder captures goal-relevant features by measuring perceptual similarity. The results for the visual metrics considered are shown in Figure~\ref{fig:img_metrics_all}.

\textbf{Familiar environment.} Visual encoders determine whether models can distinguish goal locations from visually similar scenes. A challenge evident in our doorway task (see Figure~\ref{fig:img_metrics_all}, top) where multiple similar openings must be differentiated. Effective encoders require viewpoint invariance, semantic feature recognition, and robustness to lighting changes. We assess encoding quality on goal-predicted trials using LPIPS~\cite{lpips}, DreamSim~\cite{fu2023dreamsim}, and PSNR~\cite{psnr}. GNM achieves the lowest DreamSim and highest PSNR, closely matching reference goal images. ViNT, NoMaD, NaviBridger, and CrossFormer share similar scores across all metrics, consistent with their common EfficientNet-B0 encoder. While all four recognize doorway semantics, they struggle to discriminate across visually similar rooms, leading to premature goal predictions.

\begin{table}[t]
    \centering
    \caption{Navigation metrics of \textbf{Corridor} environment with perturbations: BLUR (gray) vs SUNFLARE (yellow).}
    \renewcommand{\arraystretch}{1.3}
    \begin{tabular}{c c c c c c}  
    \toprule
    & \multicolumn{2}{c}{\textbf{Precision}} & \multicolumn{2}{c}{\textbf{Generalization}} \\
    \cmidrule(lr){2-3} \cmidrule(lr){4-5}  
        \textbf{Model} & \textbf{dist.} & \textbf{p. len.}  & \textbf{n. err.} & \textbf{g. pred.} \\
        \midrule
        
        \rowcolor{gray!5}
        \prettygnm{} & 0.62 $\pm$ 0.32 & 4.68 $\pm$ 0.35 & 0.34 $\pm$ 0.53 & 1/3  \\
        \rowcolor{gray!5}
        \prettyvint{} & 0.30 $\pm$ 0.35 & 4.70 $\pm$ 0.34 & 0.50 $\pm$ 0.67 & 2/3  \\
            
        \rowcolor{yellow!7}
        \prettygnm{} & 0.40 $\pm$ 0.15 & 4.15 $\pm$ 0.04 & 0.44 $\pm$ 0.59 & 2/3  \\
        \rowcolor{yellow!7}
        \prettyvint{} & 0.21 $\pm$ 0.04 & 4.33 $\pm$ 0.03 & 0.39 $\pm$ 0.52 & 3/3  \\
        \bottomrule
    \end{tabular}
    \label{tab:perturbation_easy_office}
\end{table}

\textbf{Out-of-distribution environment.} We evaluate model robustness in a snowy parking lot, where reflective surfaces and winter conditions introduce significant distribution shift (see Figure~\ref{fig:img_metrics_all}, bottom). As with the familiar environments, GNM maintains the lowest LPIPS and DreamSim scores. Surprisingly, ViNT, NoMaD, and NaviBridger improve over their indoor performance, suggesting that outdoor environments with more distinct spatial structure reduce the visual ambiguity that hinders indoor navigation. However, weather and lighting variations may have influenced metric values. To rigorously quantify robustness, we complement this analysis with controlled visual perturbation experiments. 




\subsection{Generalization Under Distribution Shift} 


Out-of-distribution evaluation reveals that architectural complexity does not guarantee robustness. Diffusion-based models prove fragile under distribution shift. The models GNM and ViNT achieve reliable navigation under visual perturbation (see Table~\ref{tab:perturbation_easy_office}). In snow terrain, GNM achieves the lowest goal distance and strongest visual metrics (see Figure~\ref{fig:img_metrics_all}, bottom), showing accurate goal prediction in novel conditions. NaviBridger's learned prior enables closer trajectory following (see Figure~\ref{fig:snow_mesh}), yet ViNT and NoMaD suffer from false goal prediction, causing premature termination.

\section{Lessons learned}

Our analysis reveals critical insights into the capabilities and limitations of VNMs. Our conclusions are detailed below.


\textbf{Simpler architectures may be undervalued.} Surprisingly, simpler architectures \textit{match} Transformer and diffusion-based models across several metrics. GNM, despite its simple MobileNetV2 encoder, consistently matches or slightly outperforms complex models on perceptual metrics (\textit{LPIPS, DreamSim}). Particularly, GNM closely follows ViNT performance in the Corridor and Stairs settings while, outperforming ViNT in the novel Snow setting. 



\textbf{Data is not enough.} We believe this matching between GNM and other models reflects data insufficiency rather than architectural limits. While Transformer and diffusion-based models likely possess greater representational capacity, training data has not scaled with architectural complexity. GNM and ViNT train on 70 and 80 hours respectively~\cite{gnm, vint}, and subsequent models building on these foundations increase complexity without enough additional data, likely preventing them from reaching their full potential.


\textbf{Training data composition matters.} The collision failures suggest training datasets lack sufficient obstacle and recovery examples, preventing models from learning geometric spatial reasoning. Predicted trajectories may appear plausible yet remain physically unsafe, motivating supplementary collision-avoidance mechanisms~\cite{care, zeng2025navidiffusorcostguideddiffusionmodel, nayak2025metricnet}. Future progress requires data covering rare scenarios, recovery attempts, and collision effects.

\textbf{Repeated features confuse goal prediction.} Perceptual metrics surprisingly improve in Snow over indoor settings, suggesting repetitive indoor features challenge encodes more than weather induced distribution shift. However, this finding requires careful interpretation given confounding factors from weather variations and lighting changes. 


\section{Conclusion}

We evaluate vision-based navigation models (VNMs) across diverse settings and embodiments using combined robotics and vision metrics. Despite the complex architecture of models, they frequently fail at collision avoidance and accurate goal prediction. We will release our evaluation dataset to support future benchmarking of VNMs across visually repetitive, loop closure, and novel environments. Our evaluation has two key limitations: real-world deployment constraints restrict trial counts, and, absence of social navigation. Future work should integrate geometric reasoning and explore test-time adaptation to improve robustness during deployment.



\bibliography{references}
\bibliographystyle{IEEEtran}

\end{document}